\documentclass[final,12pt]{alt2024} 

\title[Bandit Algorithms for MDPs]{Slowly Changing Adversarial Bandit Algorithms are Efficient for Discounted MDPs}
\usepackage{times}

\altauthor{%
 \Name{Ian A. Kash} \Email{iankash@uic.edu}\\
 \addr Computer Science, University of Illinois at Chicago, Chicago, Illinois, USA
 \AND
 \Name{Lev Reyzin} \Email{lreyzin@uic.edu}\\
 \addr Mathematics, Statistics, and Computer Science, University of Illinois at Chicago, Chicago, Illinois, USA
 \AND
 \Name{Zishun Yu} \Email{zyu32@uic.edu}\\
 \addr Computer Science, University of Illinois at Chicago, Chicago, Illinois, USA
}

\usepackage{mathtools} 
\usepackage{siunitx} 
\usepackage{booktabs} 
\usepackage{tikz} 
\usepackage{chngcntr}
\usepackage{blkarray}
\newcommand{\name}{\textsc{Main}}

\newcommand{\base}{\textsc{Base}}
\newcommand{\local}{\textsc{Local}}
\newcommand{\expt}{\textsc{Exp3}}

\usepackage{amsmath, amssymb}
\usepackage{amsfonts}
\usepackage{mathtools}
\usepackage{makecell}
\usepackage{dsfont}
\usepackage{xcolor}
\usepackage{color}
\usepackage{graphicx}
\usepackage{float}
\usepackage{algorithm}
\usepackage{algpseudocode}

\usepackage{booktabs} 
\usepackage{tikz} 

\usepackage{tikz}
\usetikzlibrary{automata, positioning, arrows}
\tikzset{
->, 
node distance=2.5cm, 
every state/.style={thick, fill=gray!10}, 
initial text=$ $, 
}

\newtheorem{assumption}{Assumption}

\newcommand{\Ocal}{\mathcal{O}}

\DeclareMathOperator*{\argmax}{arg\,max}

\DeclareMathOperator{\E}{\mathbb{E}} 
\DeclareMathOperator{\mbb1}{\mathds{1}}
\DeclareMathOperator{\modulo}{mod}
\newcommand{\Fcal}{\mathcal{F}}
\newcommand{\Hcal}{\mathcal{H}}
\newcommand{\Tcal}{\Pbb}
\newcommand{\Pbb}{\mathbb{P}}
\newcommand{\Scal}{\mathcal{S}}
\newcommand{\Acal}{\mathcal{A}}

\newcommand{\B}{\Big}

\newcommand{\Reg}{\mathfrak{Regret}}
\newcommand{\A}{\mathfrak{A}}
\newcommand{\sT}{\sum_{t=1}^T}

\newcommand{\qed}{\hfill $\blacksquare$}

\makeatletter
\newtheorem*{rep@theorem}{\rep@title}
\newcommand{\newreptheorem}[2]{%
\newenvironment{rep#1}[1]{%
 \def\rep@title{#2 \ref{##1}}%
 \begin{rep@theorem}}%
 {\end{rep@theorem}}}
\makeatother

\newreptheorem{theorem}{Theorem}
\newreptheorem{lemma}{Lemma}
\newreptheorem{corollary}{Corollary}

\begin{document}

\maketitle

\begin{abstract}%
{Reinforcement learning generalizes multi-armed bandit problems with additional difficulties of a longer planning horizon and unknown transition kernel. We explore a black-box reduction from discounted infinite-horizon tabular reinforcement learning to multi-armed bandits, where, specifically, an {\it independent} bandit learner is placed in each state. 
We show that, under ergodicity and fast mixing assumptions, any {\it slowly changing} adversarial bandit algorithm achieving optimal regret in the adversarial bandit setting can also attain optimal expected regret in infinite-horizon discounted Markov decision processes, with respect to the number of rounds $T$.
Furthermore, we examine our reduction using a specific instance of the exponential-weight algorithm.
} 
\end{abstract}


{

\section{Introduction}\label{sec:intro}

Reinforcement learning (RL) and multi-armed bandits (MAB) are long-standing models for decision-making problems. RL generalizes bandits with a long-term planning horizon and unknown transition dynamics. Due to these additional complexities, RL is typically viewed as a more challenging problem compared to MAB.
However, there is a large literature~\citep{kearns2002near, osband2013more, dann2017unifying, osband2017posterior, agrawal2017optimistic, fruit2018near,  jin2018q, dann2019policy, simchowitz2019non, russo2019worst, zhang2019regret, zhang2020almost, cai2020provably, zhang2020almost, neu2020unifying, pacchiano2021towards, menard2021ucb, li2021breaking, zhang2021reinforcement} that guarantees RL can achieve the optimal regret $\Omega(\sqrt{T})$ in the dependency of the number of rounds $T$, and is often optimal in terms of the cardinalities, $S$ and $A$, of state and action spaces.
Recent episodic horizon-free works \citep{wang2020long, zhang2021reinforcement, zhang2022horizon, li2023horizon} further show the potential to close the formal complexity gap between RL and bandits, with RL's regret approaching the lower bound of the (contextual) MAB problem $\Omega(\sqrt{SAT})$ \citep{bubeck2013bounded, auer1995gambling, gerchinovitz2016refined}. These findings imply that the longer planning horizon and unknown transition kernels in RL may not introduce additional difficulties compared to bandits.

We therefore ask: {\it Is there a reduction from tabular reinforcement learning to multi-armed bandits?} Specifically, in a {\it decentralized} setting, could one place an {\it independent} bandit learner in each state (referred as local learners), such that this set of local learners achieves sub-linear regret in MDPs collectively, without needing to acquire information (for example value estimations) from their co-learners, except for the shared global rewards?

We answer this question positively for discounted infinite-horizon MDPs. We prove that, under ergodicity and fast mixing assumptions, one could trivially place $\tilde{\mathcal{O}}(S)$\footnote{$\tilde{\mathcal{O}}(\cdot)$ compresses polylog dependencies.} arbitrary {\it slowly changing} bandit algorithms to achieve a regret bound of $\tilde{\mathcal{O}} (\mathrm{poly}(S, A, H, \tau, \tfrac{1}{\beta}, \tfrac{1}{1-\gamma}) \cdot (\sqrt{T} + c_T T))$ (which depends on various problem parameters specified in later sections), if the bandit learners are optimal in the adversarial bandit setting.
Here, $c_T$ represents the changing rate for the chosen bandit algorithm.
The regret bound is optimal with respect to $T$ (up to polylogarithmic factors) when $c_T$ is $\tilde{\mathcal{O}}(1/\sqrt{T})$,
which is a mild requirement as discussed in later sections.

Despite the decentralized framework where each state is managed by an independent learner being a compelling problem in itself, the decoupling from the temporal difference framework makes it straightforward to leverage techniques from the bandit toolbox. 
For instance, in Section~\ref{sec:regret} we show how our reduction framework effectively handles delayed feedback, benefiting from the robustness of adversarial bandits to such feedback. 
This also opens up possibilities for straightforward translation of existing bandit results, such as delayed or aggregated feedback~\citep{joulani2013online, pike2018bandits}, to MDPs, especially since these settings are gaining traction in RL as well~\citep{howson2021delayed, jin2022near, mondal2023reinforcement}.
In addition, understanding the reduction to independent learners can be connected to multi-agent RL, where such decentralization allows mitigating the curse of multiagency~\citep{jin2022v, cravic2023decentralized}, and can be also bridged to Monte Carlo methods, as detailed in Section~\ref{sec:related}.

\section{Related Work}\label{sec:related}

The work most closely related to ours is perhaps that of~\citet{cheng2020reduction}, who propose a reduction from RL to continuous online learning~\citep{cheng2020online} under a generative oracle setting. This setting allows algorithms to query transitions from the true dynamics without interacting with the environment. In addition to the generative model requirement, our work is significantly different from theirs in the sense that their work considers centralized no-regret learners, communicating through the value function estimations while ours is decentralized.

On the other hand, diverging from the canonical temporal difference scheme makes the Monte Carlo evaluation a natural choice for our reduction, as detailed in Section~\ref{sec:algo}. This positions our work within the realm of Monte Carlo methods: for example, Monte Carlo Exploring Starts (MCES)~\citep{sutton2018reinforcement}. Similar to MCES, our reduction associates each state with an independent decision-maker using Monte Carlo estimations.
The primary difference lies in the exploration technique: MCES uses exploring starts\footnote{Exploring starts sample an initial $(s_0, a_0)$ randomly for each episode, ensuring all $(s, a)$ are visited infinitely often.}, whereas in our reduction, exploration is partly delegated to the bandit learners.
Despite being considered as ``one of the most fundamental open theoretical questions in reinforcement learning''~\citep{sutton2018reinforcement}, 
MCES had relatively few guarantees until recent works on its convergence~\citep{wang2021convergence, liu2021convergence, dong2022convergence, winnicki2023convergence}, while an earlier result by~\citet{tsitsiklis2002convergence} requires more restrictive assumptions.

In terms of implementations and technical tools, our reduction aligns more closely with research in the area of {online MDPs}~\citep{even2009online, gergely2010online, rosenberg2019online, jin2020learning}. Specifically, \citet{even2009online} implemented a framework where each state is managed by an expert algorithm, while \citet{gergely2010online} proposed a model with a bandit learner assigned to each state.
Yet, in the work on online MDPs, policy evaluation is still done in a temporal difference fashion, which differs from our reduction.
The ``slowly changing'' property required by our reduction, is also an important insight from these works. One can analyze slowly changing policies with their stationary distributions which are generally easier to handle, see Lemma~\ref{lemma:mixing-error} for details. But in general our analysis is still very different because of our decentralized setup. We further leveraged the slowly changing property to give our results in Section~\ref{sec:obj-mismatch} and Section~\ref{sec:sticky-bandit} to address the corresponding difficulties raised by such decentralization.

{\bf Additional Related Works.}
We consider infinite-horizon discounted MDPs, akin to the setting considered by the reduction in~\citet{cheng2020reduction}. We would like to note that one could often translate the results from infinite horizon setting to episodic setting but not vice versa~\citep{ortner2020regret,dong2019q}, because of infinite planning horizon and lack of restarting mechanism. In contrast to episodic MDPs listed in Section~\ref{sec:intro},
the study in the area of infinite-horizon discounted MDPs~\citep[][etc.]{dong2019q, liu2020regret, he2020nearly, zhou2021provably, yang2021q, yan2023efficacy} 
is relatively limited.
In terms of planning horizon, in addition we have the category of infinite horizon average reward setting~\citep[][etc.]{auer2008near, ouyang2017learning, talebi2018variance,fruit2018efficient, ortner2020regret, dewanto2020average, wei2021learning, zhang2023sharper}.
While the majority of works discussed above are measured by regret, the line of works~\citep[][etc.]{kakade2003sample, strehl2006pac, strehl2008analysis, kolter2009near, bartlett2012regal, szita2010model, lattimore2012pac, lattimore2013sample, dann2015sample, modi2020sample, xu2020improving} that established with {\it sample complexity of exploration}~\citep{kakade2003sample} is also a major direction.

}

\section{Preliminaries}

\subsection{Discounted Infinite Horizon MDPs}

A tabular MDP $\mathcal{M}$ is often described by a 5-tuple $(\Scal, \Acal, \mathbb{P}, r, \gamma)$, where $\Scal$ and $\Acal$ are finite state and action spaces, respectively. We denote their cardinality by $S \coloneqq |\Scal|$ and $A \coloneqq |\Acal|$. Let $\Delta(X)$ be all probability distributions over space $X$, $\mathbb{P}: \Scal \times \Acal \to \Delta(\Scal)$ is the \textit{unknown} stochastic transition function, $r: \Scal \times \Acal \to \Delta([0, 1])$ is the \textit{unknown} reward function, and $\gamma \in [0, 1)$ is a discount factor. 

\noindent\textbf{Policy.} A policy is a mapping $\pi: \Scal \to \Delta(\Acal)$. In our case, a policy $\pi_t$, at time $t$, is collectively determined by the set of bandit learners, as each learner determines the strategy for a state $\pi_t(\cdot|s)$, see Algorithm~\ref{alg:main} for details.

\noindent\textbf{Value Functions.} Given a policy $\pi$, the state value function $V^\pi(s)$ and state-action value function $Q^\pi(s, a)$ are defined as,
\begin{align*}
    V^\pi(s) \coloneqq \E \left[ \sum_{t=0}^\infty \gamma^t R_t|\pi, S_0 = s \right], \quad
    Q^\pi(s, a) \coloneqq \E \left[ \sum_{t=0}^\infty \gamma^t R_t|\pi, S_0 = s, A_0 = a \right].
\end{align*}

\noindent\textbf{Optimality.} The optimal policy $\pi^* \! \coloneqq \! \argmax_\pi V^\pi(s)$, for all $s \in \Scal$. 
$V^{*}, Q^{*}$ denote value functions corresponding to $\pi^*$.

\noindent\textbf{State Distributions.} The state distribution $\nu$ at $t+1$ is recursively characterized by $\nu_{t+1} \coloneqq \nu_{t} \Pbb^{\pi_t}$, where $\Pbb^{\pi}$ is the transition kernel induced by $\pi$ and we denote $\nu_1$ as the initial distribution. The stationary distribution $\mu$ of a policy $\pi$ is the left eigenvector of $\Pbb^{\pi}$, i.e. $\mu^{\pi} \Pbb^{\pi} = \mu^{\pi}$. 

For brevity, we use $Q_t$, $V_t$, $\mu_t$ to denote $Q^{\pi_t}$, $V^{\pi_t}$, $\mu^{\pi_t}$, respectively.

\subsection{Regret}\label{sec:regret}
An obstacle to address is the different languages used in bandits and infinite-horizon discounted RL literature. While bandits community often measures algorithms' performance by regret, the community of infinite-horizon RL often uses the \textit{sample complexity of exploration}~\citep{kakade2003sample} (sample complexity in short). These two notions are often not translatable to each other, as regret measures the quantity of cumulative sub-optimailities but sample complexity counts the number of sub-optimailities that violate a threshold $\epsilon$. 
In addition to the difference between cumulative sub-optimality value vs. number of sub-optimailities, the sample complexity is not a function of the total number of rounds $T$. 
Two $T$-step optimal MDP learners, in the sense that they reach the optimal policy in $T$ steps, could be considered equally ``good'' in terms of sample complexity during those initial $T$ steps, but they could show significant differences in terms of regret measures.

To the end of a black-box reduction, we align the performance measures by leveraging a recent regret definition for discounted infinite-horizon MDPs, used by~\citet{liu2020regret, he2020nearly, zhou2021provably}, which measures the cumulative sub-optimality $V^*(s_t) - V_t(s_t)$ that defined by the state value function.
\begin{definition} Regret for infinite-horizon discounted MDPs
\label{def:regret}
\begin{align*}
    \Reg(T) &\coloneqq \sT \B[ V^*(s_t) - V_t(s_t) \B].
\end{align*}
\end{definition}
While this regret and sample complexity are not directly comparable (for example, a policy with fewer, yet larger suboptimalities may have worse regret but better sample complexity, or vice versa), bounds on sample complexity can however imply upper bounds on regret. \citet{he2020nearly} shows that a sample complexity bound of $\Ocal(M\epsilon^{-\alpha})$ implies a maximum regret of $\Ocal(M^{1/(\alpha+1)}(1-\gamma)^{-1/(\alpha+1)}T^{\alpha/{\alpha+1}})$. This suggests that, for instance, a ${\Ocal}(\epsilon^{-2})$ sample complexity implies a worst-case regret of ${\Ocal}(T^{2/3})$. Although quantifying the tightness of this approximate translation is challenging, it offers a general sense of the regret notion's strength. Further insights into the comparison between sample complexity and this regret notion are discussed in~\citet{liu2020regret}.

\subsection{Assumptions}\label{sec:assumptions}

We make two additional assumptions.

\begin{assumption}\label{asm:uniform-bound}
   The stationary distributions are uniformly bounded away from zero. $$\inf_{\pi, s}\mu^\pi(s) \geq \beta\ \ \mathrm{for\ some}\ \beta > 0.$$ 
\end{assumption}

\begin{assumption}\label{asm:fast-mixing}
    There exists some fixed positive $\tau$ such that for any two arbitrary distributions $d$ and $d'$ over $\Scal$,
    \begin{equation*}
        \sup_\pi||(d-d')\Pbb^\pi||_1 \leq e^{-1/\tau}||d-d'||_1 ,
    \end{equation*}
    where $\tau$ is the mixing time, we further assume $\tau \geq 1$ without loss of generality.
\end{assumption}

Assumption~\ref{asm:fast-mixing} bounds the mixing time, of Markov chain induced by some policy $\pi$, by $\tau$. 
It also implies the existence and uniqueness of stationary distribution $\mu^\pi$.
{ These assumptions combined guarantee the MDP is ``well behaved" in the sense that all states are likely to be visited often, thus ensuring frequent updates for each bandit learner, regardless of the policy and starting point.
This is essential as an ``out-dated" bandit would potentially hurt the overall performance.
In addition, our assumptions play a similar role to the exploring starts in MCES, as it ensures exploration over $\Scal$, akin to the exploration over $\Scal\times\Acal$ provided by exploring starts.
}
These assumptions have been used in prior work on online learning in MDPs such as that of \citet{gergely2010online, rosenberg2019online}, and the latter is also made in literature of stochastic games such as~\citet{etesami2022learning}.\footnote{Our setting is akin to cooperative games to certain extend, in the sense that local learners aimed to maximize shared global payoff without knowing the strategy of its co-learners.}
For scenarios without these assumptions, a counter-example is provided in Appendix~\ref{appendix:counter-example}.

\subsection{Slowly Changing Algorithms}

Our main result requires that the bandits placed in each state are slowly changing, for which we now provide a formal definition. To measure the change rate of an algorithm, we first introduce the 1-$\infty$ norm. For a ``conditional matrix'' $\mathbf{M}(y|x)$, it is defined as $||\mathbf{M}||_{1, \infty} \coloneqq \max_x \sum_y |\mathbf{M}(y|x)|$, which can be used to measure the difference between two policies $||\pi - \pi'||_{1, \infty} = \max_s \sum_a [\pi(a|s) - \pi'(a|s)]$.
\begin{definition}[Slowly Changing]\label{def:slow}
An algorithm $\A$ is slowly changing with a (non-increasing) rate of $c_T$ if, for all $t$,
$ ||\pi_{t+1} - \pi_t||_{1, \infty} \leq c_T$,
where $\pi_t$ is the policy produced by $\A$ at time $t$.
\end{definition}
Note that, throughout this paper, we assume the number of rounds $T$ is known. When $T$ is unknown, it can be managed using the standard doubling trick, see~\citet{shalev2012online} for example.

Our analysis relies on using bandits in our algorithm that themselves are slowly changing. This slowly changing definition also applies to bandit algorithms, as one could consider the state space of bandit learners as a singleton $\{s\}$.
The assumption of slowly changing bandits is mild and has been used in prior works on online learning in MDPs, such as~\citet{even2009online,gergely2010online}. For completeness, we prove in Section~\ref{sec:exp3} that \expt~\citep{auer2002nonstochastic} is slowly changing in this respect, a fact also observed and indirectly used by \cite{gergely2010online}.

\section{A Black-Box Algorithm}\label{sec:algo}

We now present our framework in Algorithm~\ref{alg:main}, which is based on a slowly changing bandit algorithm, referred to as \local. Accordingly, Algorithm~\ref{alg:main} is named \name. The key idea of our reduction is to deploy an instance of \local~in each state, thereby determining the strategy for that particular state.

Furthermore, we require that this bandit algorithm can accommodate delayed feedback. Robustness to delays allows us to wait to provide feedback to the algorithm, until a time such that the difference between the return at that time and the return of the full trajectory is sufficiently small, ensuring the return estimation is sufficiently accurate for the corresponding action pulled. 
We discuss how delayed feedback can be addressed in a black-box fashion in Section~\ref{sec:delayed-bandits}. 
In addition, as bandits may be updated over the course of the trajectory, the slowly changing property guarantees these changes have only a ``small'' effect on the expected return. 
Combined, these properties ensure that error in the feedback used to update the bandit, relative to 
the true value, 
is manageable.

\begin{algorithm2e}[!t]
\caption{ (\name) Bandits for MDPs}\label{alg:main}
\DontPrintSemicolon  

\textbf{Require:} $\gamma \in [0, 1), T, H\coloneqq \left\lceil \log_\gamma{ \frac{1-\gamma}{\sqrt{T} } } \right\rceil$, \local

\textbf{Initialize:} \{\local$_s: s=1, 2, \cdots, |\Scal|$\}
\Comment{Initialize one instance for each state.}

\For{$t = 1, 2, \dots, T$}{
    Observe state $S_t$\;
    Obtain action distr. $\pi_t(\cdot|S_t)$ (from $\local_{S_t}$) \;
    Draw $A_t \sim \pi_t(\cdot|S_t)$\;
    Observe reward $R_t$\;    
    \If{$t > H$}{
    Cumulative gain $\bar{G}_{t-H} = \sum_{i=t-H}^{t} \gamma^i R_i$\;
    Return $\bar{G}_{t-H}$ to \local$_{S_{\{t-H\}}}$ as feedback \Comment{Delayed feedback and local update.}
    }
}
\end{algorithm2e}

\subsection{Monte Carlo Estimator}\label{sec:estimator}
To estimate the value of a policy, one could straightforwardly use a Monte Carlo estimator $G_t \coloneqq \sum_{i=t}^\infty \gamma^{i-t} R_i$, as implemented in methods such as REINFORCE and MCES~\citep{sutton2018reinforcement}. In our setting of infinite horizon MDPs, we practically use its finite horizon counterpart $\bar{G}_t \coloneqq \sum_{i=t}^{t+H} \gamma^{i-t} R_i$, with the 
{ {\it effective horizon} $H = \mathcal{O}(\log \sqrt{T} / \log(1/\gamma)) = \tilde{O}(1/\log(1/\gamma))$}, 
as defined in Algorithm~\ref{alg:main}.
However, given that our policy changes due to local bandit updates during the period of collecting $G_t$, $G_t$ is not an unbiased estimator of $Q_t(S_t, A_t)$. This issue also applies to $\bar{G}_t$. Instead, $G_t$ and $\bar{G}_t$ are unbiased to the conditional expectations below,
\begin{align*}
    U_t(s, a) \coloneqq \E[G_t|S_t = s, A_t = a, \Fcal_{t-1}], \quad
    \bar{U}_t(s, a) \coloneqq \E[\bar{G}_t|S_t = s, A_t = a, \Fcal_{t-1}].
\end{align*}
Note that $U_t$ is a non-stationary analogue of action-value function $Q_t$. The difference is that $Q_t(s, a)$ depends only on the stationary policy $\pi_t$ while $U_t$ depends on the past histories $\Fcal_{t-1} \coloneqq \{(S_i, A_i, R_i): 1 \leq i \leq t-1\}$, in additional to the MDP. As with $Q_t$, $U_t$ is well defined even at states and actions other than those visited at time $t$.

\section{Regret Analysis}\label{sec:regret}

In developing the proof for our main theorem, Theorem~\ref{thm:main}, we (1) begin by decomposing the global regret into local regrets; (2) then address the challenges posed by our algorithm designs and the regret decomposition; (3) and conclude the final theorem with prior results.

\subsection{Global to Local}

We begin by defining \textit{local regret with oracle feedback} (referred to as {\it local regret} when no confusion arises) as follows,
\begin{definition}\label{def:local-regret}
    For $s \in \Scal$, the local regret with oracle feedback $Q_t$ is defined as:
    \begin{align*}
        \mathfrak{R}_{s}(T) \coloneqq \sT \B[ \E_{a \sim \pi^*(s)} Q_t(s, a)  - \E_{a \sim \pi_t(s)}  Q_t(s, a) \B]
        = \sT \sum_{a \in \Acal} \left( \pi^*(a|s) - \pi_t(a|s) \right) Q_t(s, a) .
    \end{align*}
\end{definition}

We adapt the idea by~\citet{even2009online} that the global regret can be decomposed into local ones, to our discounted setting along with the new (global) regret definition.
In Lemma~\ref{lemma:decomposition}, we show that the expected regret of learning in MDPs can be bounded by the cumulative regret of the set of local bandit problems, assuming the feedback $Q$-functions are given by an oracle. This can be done with the help of \textit{performance difference lemma}~\citep{kakade2002approximately, kakade2003sample}, which is deferred to Appendix~\ref{sec:decomposition} along with the proof of Lemma~\ref{lemma:decomposition}.
\begin{lemma}\label{lemma:decomposition}
    The global $\Reg(T)$ can be bounded by the cumulative local regret, scaled by $\frac{1}{1-\gamma}$ 
    \begin{align*}
        \Reg(T) &= \sT \left( V^*(s_t) - V^{\pi_t}(s_t) \right) \leq \frac{1}{1-\gamma} \sum_{s \in \Scal} \mathfrak{R}_{s}(T) .
    \end{align*}
\end{lemma}

Now we decompose our problem into smaller pieces, where each state is in fact corresponding to a \local\ bandit learner. This decomposition allows us to conduct analysis at the bandit level.

\subsection{Objective Mismatch}\label{sec:obj-mismatch}

While Lemma~\ref{lemma:decomposition} helps us to break down our problem into sub-problems, it also introduces some challenges. The first major challenge is the discrepancy between the oracle feedback $Q_t$ and our approximation target $\bar{U}_t$. 
As discussed in Section~\ref{sec:estimator}, $\bar{G}_t$ is an unbiased estimator of $\bar{U}_t$ but is biased to $Q_t$, while the local regret $\mathfrak{R}_s$ is unfortunately measured using $Q_t$. 
Therefore, we refer to this issue as objective mismatch.

To address the mismatch between objectives, we rely on the slowly changing property. Intuitively speaking, the deviation of expected return, $\bar{U}_t$ versus $Q_t$, should be relatively small if the policy changes sufficiently slow. 
Thanks to the slowly changing guarantee, we show that one could bound the gap between $\bar{U}_t$ and $Q_t$ in Lemma~\ref{lemma:approx-error}. 

\begin{lemma}\label{lemma:approx-error}
    If \name~is slowly changing with a rate of $c_T$, then 
    \begin{equation*}
        \left| \bar{U}_t(s, a) - Q_t(s, a) \right| \leq \frac{H(S+HA)}{1-\gamma} c_T + \frac{1}{\sqrt{T}}.
    \end{equation*}
\end{lemma}

This gap shows that the additional error introduced by the non-stationarity during the effective horizon $H$ can be controlled. We defer its proof to Appendix~\ref{appendix:obj-mismatch} as it is quite technical.

\begin{corollary}\label{coroll:obj-mismatch}
    Let $\mathfrak{R}_s^{\bar{U}}(T) \coloneqq \sT \sum_{a \in \Acal} (\pi^*(a|s) - \pi_t(a|s) ) \bar{U}_t(s, a)$, we have
    $$\left| \mathfrak{R}_s^{\bar{U}}(T) - \mathfrak{R}_s(T) \right| \leq \frac{2H(S+HA)}{1-\gamma} c_T T + 2\sqrt{T}. $$
\end{corollary}

Corollary~\ref{coroll:obj-mismatch} shows that the difference between local regret measured with $\bar{U}_t$ and local regret with oracle feedback $Q_t$ is manageable, if $c_T$ is sufficiently small. 
It hence allows us to analyze the local problems using the oracle feedback $Q_t$, instead of $\bar{U}_t$ that the actual feedback $\bar{G}_t$ approximates. This largely simplifies our subsequent analysis as $Q_t$ is much easier to handle.

\subsection{Sticky Bandits}\label{sec:sticky-bandit}

Another challenge is that at each time we are only in a single state so only a single bandit is updated, while local regret $\mathfrak{R}_s(T)$ is measured over the entire time span $T$.  We term this the {\em sticky} bandit setting, in the spirit of sticky actions in the Arcade learning environment~\citep{machado2018revisiting}, because from the perspective of a bandit it is given feedback and the opportunity to change its policy only occasionally. 

\subsubsection{General Definitions}\label{sec:sticky-bandits-def}
We start with a general definition to isolate the issue of sticky bandits.

\begin{definition}[Sticky Bandit]
    Let $T$ be the total number of rounds, and $X_i$ be the time $t$ at which the bandit is allowed to act for the $i$-th time.
    The action is sticky in the sense that $p_t(a) = p_{X_i}(a)$ for $X_i \leq t < X_{i+1}$, where $p_t$ is the distribution over $\Acal$ at time $t$.
\end{definition}

As the decomposition lemma requires the regret of a local bandit during the full time span $T$, we thereby define three regret notions, full (time) span regret, observed regret and unobserved regret.
\begin{definition}
    Full-span regret $R^{\text{fs}}(T)$, observed regret $R^{\text{ob}}(T)$ and unobserved regret $R^{\text{un}}(T)$
    \begin{align*}
        R^{\text{fs}}(T) &\coloneqq \sT \sum_{a \in \Acal} \left( p^*_{t}(a) - p_{t}(a) \right) r_t(a) \\
        R^{\text{ob}}(T) &\coloneqq \E_{\{X_i\}} \sum_{t\in\{X_i\}} \sum_{a \in \Acal} \left( p^*_{t}(a) - p_{t}(a) \right) r_t(a) \\
        R^{\text{un}}(T) &\coloneqq R^{\text{fs}}(T) - R^{\text{ob}}(T) .
    \end{align*}
\end{definition}

It is note-worthy that $R^{\text{fs}}(T)$ degenerates to local regret $\mathfrak{R}_{s}(T)$, if one apply $p_t(a)=\pi_t(a|s)$, $p^*_t = \pi^*(a|s)$ and $r_t(a)=Q_t(s, a)$. Therefore, if one could prove that sub-linear observed regret implies sub-linear full-span regret, then we could translate observed regret of local bandits to global regret in MDPs. 
Assumptions made in Section~\ref{sec:assumptions} ensure that each state will be visited sufficiently often, 
meaning each local bandit will be updated often.
%
However, it is also generally impossible for an arbitrary bandit algorithm to be no-regret, for the full time span, with these assumptions alone. 

%

\noindent\textbf{A Hard Instance for Sticky Bandits.} Consider a sticky and adversary setting with two actions $a_1$ and $a_2$. We assume the bandit learner is only able to pull every 10 rounds and the first pull is at $t=1$ without loss generality. 
And the adversary choose the reward function below 
$$
r_{a_1}(t) = 
\begin{cases}
1 &\text{$t \% 10 \in [1, 5]$}, \\
0 &\text{otherwise}.
\end{cases}\quad
r_{a_2}(t) = 
\begin{cases}
0 &\text{$t \% 10 \in [1, 5]$},\\
1 &\text{otherwise}.
\end{cases}
$$

Then a bandit learner is likely leaning to pull $a_1$, as it is never able to observe that $a_2$ achieves a reward of $1$. It in turn implies that the bandit player will have an $\Ocal(T)$ full-span regret.  
This challenge is caused by the possibility of dramatic reward changes. Therefore one could not predict what is the regret while the bandit player cannot pull and observe, even if it pulls frequently enough. However, in Section~\ref{sec:sticky-mdps} we prove that the reward/feedback function of local bandits are also in the family of slowly changing functions. Therefore one could estimate the regret occurred, when bandits are not able to react, by its latest regret seen. %

\subsubsection{Learning in MDPs}\label{sec:sticky-mdps}

We now connect these regret definitions to learning in MDPs, 
by applying  $\pi^*$ as the comparator and $Q_t$ as the feedback function.

\begin{definition}
    Full-span regret, observed and unobserved regret in MDPs are defined as follows
    \begin{align*}
        R^{\text{fs-mdp}} &\coloneqq \sum_{s\in\Scal} \sT \sum_{a\in\Acal} (\pi^*(a|s) - \pi_t(a|s))Q_t(s, a) 
        \eqqcolon \sum_{s\in\Scal} \mathfrak{R}_s(T) \geq (1-\gamma)\Reg(T) \\
        R^{\text{ob-mdp}} &\coloneqq \sum_{s\in\Scal} \sT \nu_t(s) 
        \sum_{a\in\Acal} (\pi^*(a|s) - \pi_t(a|s))Q_t(s, a) \quad 
        R^{\text{un-mdp}}\coloneqq R^{\text{fs-mdp}} -  R^{\text{ob-mdp}}
    \end{align*}
    where the inequality follows from Lemma~\ref{lemma:decomposition}, { and $\nu_{t} \coloneqq \nu_{t-1} \Pbb^{\pi_{t-1}}$ denotes the state distribution at $t$.}
\end{definition}
In MDPs, the full-span regret $R^{\text{fs-mdp}}$ is simply defined by accumulating all local regret $\mathfrak{R}_{s}(T)$, given the aforementioned choices of comparator and feedback function. The observed regret $R^{\text{ob-mdp}}$ similarly accumulates the observed local ones, based on the state visitation distribution $\nu_t$.

It is clear that the observed regret is sub-linear if the bandit learners are no-regret. However, this conclusion is not sufficient to help us infer anything about full-span regret. As discussed in Section~\ref{sec:sticky-bandits-def}, the first challenge is the potential dramatic change of feedback, which in turn leads to difficulty to measure the unobserved regret. We show, in Lemma~\ref{lemma:delay-error-Q}, that $Q_t$ is indeed slowly changing because $\pi_t$ is, with its proof deferred to Appendix~\ref{appendix:delay-error-Q}.

\begin{lemma}\label{lemma:delay-error-Q}
    If \name~is slowly changing with a non-increasing rate of $c_T$, we have
    \begin{align*}
        |Q_{t+n}(s, a) - Q_t(s, a)| &\leq \frac{(S + HA)n}{1-\gamma} c_T + \frac{2}{\sqrt{T}} .
    \end{align*}
\end{lemma}

The second difficulty is raised by the state distribution $\nu_t$. It is generally difficult to analyze $\nu_t$ because it is a product of a sequence of prior policies. 
We therefore leverage the insight from the online MDPs literature~\citep{even2009online, gergely2010online} that $\nu_t$ is close to its stationary distribution $\mu_t$ if the algorithm is slowly changing, as shown in Lemma~\ref{lemma:mixing-error} whose proof can be found in Appendix~\ref{appendix:mixing-error}. It is much easier to conduct analysis with the stationary distributions.
\begin{lemma}\label{lemma:mixing-error}
    If the sequence of policies $\{ \pi_t \}$ is slowly changing with rate $c_T$, then 
    $$||\nu_t - \mu_t||_1 \leq \tau(\tau+1)c_T + 2e^{-(t-1)/\tau}.$$
\end{lemma}

\begin{corollary}\label{coroll:mixing-error} 
    As a result of Lemma~\ref{lemma:mixing-error}, one could bound the observed/unobserved regret as follows,
    \begin{align*}
        R^{\text{ob-mdp}} 
        &\leq \kappa T + 
        \underbrace{ \sum\nolimits_{s\in\Scal} \sum\nolimits_{t=1}^T \mu_t(s) \sum\nolimits_{a\in\Acal} (\pi^*(a|s) - \pi_t(a|s))Q_t(s, a) }\nolimits_{  =: \tilde{R}^{\text{ob-mdp}}}
        \\
        R^{\text{un-mdp}} 
        &\leq \kappa T + 
        \underbrace{ \sum\nolimits_{s\in\Scal} \sum\nolimits_{t=1}^T \left(1 - \mu_t(s) \right) \sum\nolimits_{a\in\Acal} (\pi^*(a|s) - \pi_t(a|s))Q_t(s, a)  }\nolimits_{   =: \tilde{R}^{\text{un-mdp}}}
        \\ 
        \kappa &= \left( \tau (\tau+1) c_T + 2e^{-(t-1)/\tau} \right) \big/ (1-\gamma).
    \end{align*}
\end{corollary}

These bounds are useful as $\mu_t$ is uniformly bounded below given Assumption~\ref{asm:uniform-bound}, which in turn implies uniformly sufficient visitation. Combined with the slowly changing feedback as established in Lemma~\ref{lemma:delay-error-Q}, these conditions together are adequate to address the challenges posed by the sticky bandit issue.
We have now converted the original problem associated with $\nu_t$, to a surrogate problem with stationary distributions $\mu_t$.

The observed and unobserved regret, $\tilde{R}^{\text{ob-mdp}}$ and $\tilde{R}^{\text{un-mdp}}$, for this surrogate problem are defined in Corollary~\ref{coroll:mixing-error}.
Bounding the surrogate unobserved regret $\tilde{R}^{\text{un-mdp}}$ leads to a bound of the original regret $R^{\text{un-mdp}}$.
Now we are ready to show, in Lemma~\ref{lemma:full-regret}, that $\tilde{R}^{\text{un-mdp}}$ can be bounded by $\tilde{R}^{\text{ob-mdp}}$ up to a factor $\beta$ as well as additional terms that are sub-linear in $T$, with proper choice of $c_T$.
\begin{lemma}\label{lemma:full-regret}
    If Assumption~\ref{asm:uniform-bound} and \ref{asm:fast-mixing} are satisfied, and the \local\ learner is slowly changing with a rate $c_T$, we have
    $$ \tilde{R}^{\text{un-mdp}}(T) \leq \frac{\tilde{R}^{\text{ob-mdp}}(T)}{\beta} + \frac{2(S+HA)}{(1-\gamma)\beta^3} c_T T + 4S\sqrt{T} .$$
\end{lemma}

It in turn leads to our second key result regarding the full-span regret $R^{\text{fs-mdp}}(T)$ in Corollary~\ref{coroll:sticky-bandits}, following from Corollary~\ref{coroll:mixing-error} and Lemma~\ref{lemma:full-regret}.
{
\begin{corollary}\label{coroll:sticky-bandits}
    Suppose assumption~\ref{asm:uniform-bound} and \ref{asm:fast-mixing} are satisfied. If a slowly changing \local~bandit, with a rate of $c_T$, enjoys $R^{\text{ob-mdp}}(T) = \tilde{\Ocal}(g(\cdot) S \sqrt{T})$ observed regret,
    then the full-span regret is
    $$R^{\text{fs-mdp}}(T) =  \tilde{\Ocal} \left( \frac{g(\cdot)S}{\beta}\sqrt{T} + \left( \frac{S+HA}{(1-\gamma)\beta^3} + \frac{\tau^2}{1-\gamma} \right) c_T T \right) $$
    where $g(\cdot)$ is a function of other problem parameters, such as $A$ and $H$, specified in later sections. 
\end{corollary}
}

\subsection{Delayed Feedback}\label{sec:delayed-bandits}

Due to our construction, we introduced {\it constant} feedback delays into our Algorithm~\ref{alg:main}. For the purpose of black-box reduction, one need to address the delays in a black-box fashion. We leverage the result from~\citet{joulani2013online}, which bounds the regret of delayed problems for arbitrary bandit algorithm with its non-delayed guarantees. They provide a black-box algorithm for (arbitrary) delay. The algorithm is presented in Algorithm~\ref{alg:bold}, in the context of constant delay. However, this step may not be necessary in practice, as many adversary bandit algorithms have been shown robust to constant delay~\citep{gergely2010online, joulani2013online, cesa2016delay, pike2018bandits, bistritz2019online, thune2019nonstochastic}, etc.
See further discussion in Section~\ref{sec:exp3-delay}.

\begin{algorithm2e}[!t]
\caption{($\local_s$) black-box online learning under (constant) delayed feedback}\label{alg:bold}
\DontPrintSemicolon  

\textbf{Require:} constant delay $H$\;
\textbf{Initialize:} {$\base^h_s: h=1,2,\dots,H+1$}\;

\For{$t=1,2, \dots, T$} 
    {Set $h_t = [t \text{~mod~} (H+1)] + 1 $\;
    Choose $\base^{h_t}_{s_t}$ to make prediction\;
    \If{$t > H$}{
        Receive feedback $\bar{G}_{t-H}$\; %
        Update $\base^{h_{t-H}}_{s_{t-H}}$ with $\bar{G}_{t-H}$\; %
        }
    }
\end{algorithm2e}

The essence of the construction is using $H+1$ \base~instances, so that each instance can update after receiving the feedback of its last decision. Therefore, a delayed problem is reduced to $H+1$ non-delayed problems. Now it is possible to handle the delayed feedback in a black-box fashion.

\begin{lemma}\label{lemma:bold}
    [\cite{joulani2013online}] Suppose that the \base\ used in \local\ enjoys an expected regret bound $R^{\base}(T)$ in non-delayed setting. Assume, furthermore, that the delays are constant $H$. Then the expected regret of \local\ after $T$ time steps satisfies
    \begin{align*}
        R^{\text{\local}} \leq (H+1)R^\base\left( T/(H+1) \right) .
    \end{align*}
\end{lemma}

{
\begin{corollary}\label{coroll:delayed-bandits}
    Suppose \base\ has $\tilde{\Ocal}({ f(A)}\sqrt{T})$ regret, Algorithm~\ref{alg:bold} then enjoys  
    $\tilde{\Ocal}( { f(A)} \sqrt{HT} )$ regret, 
    where $f(A)$ denotes the dependency on $A$.
\end{corollary}
}

\subsection{\name~is Slowly Changing}
To summarize, the black-box reduction flow is now 
$$\name \xrightarrow{\text{Algo.~\ref{alg:main}}} \local_s \xrightarrow{\text{Algo.~\ref{alg:bold}}} \base_s^h $$
with $\Ocal(H)$ bandit learners per state, and $\Ocal(HS)$ in total.

While Section~\ref{sec:obj-mismatch} and Section~\ref{sec:sticky-bandit} rely on the slowly changing property of \name, we have not yet show that \name~is slowly changing if \base~is slowly changing. It is not difficult to see that \name~is slowly changing if \local~is. The corresponding lemma and its proof can be found in Appendix~\ref{appendix:slowly-changing}. 
Unfortunately, even if \base~is slowly changing, $\local$ is not necessarily slowly changing due to the switching mechanism - alternating among various \base~instances - designed by Algorithm~\ref{alg:bold}, as each of the $H+1$ \base~bandits could have arbitrarily different policies.

However, 
one could preserve the slowly changing property by incorporating the timestep $h$ as part of the state. In other words, one could augment the state space $\Scal$ by concatenating a state $s$ with a time stamp $h \in \Hcal \coloneqq \{1, 2, \dots, H+1\}$.
Definition~\ref{def:H-MDP} gives a formal statement of $\Hcal$-augmented MDPs. Similarly, constructing timestep (of episodes) as part of the state is often seen in episodic settings~\citep[][etc.]{jin2018q, wang2021convergence}. 

\begin{definition}\label{def:H-MDP}
    Given a MDP $\mathcal{M}=(\Scal, \Acal, \mathbb{P}, r, \gamma)$, and let $\Hcal \coloneqq \{1, 2, \dots, H+1\}$. We define the $\Hcal$-argumented MDP as  $\tilde{M} = (\tilde{\Scal}, \Acal, \tilde{\mathbb{P}}, \tilde{r}, \gamma)$, where $\tilde{\Scal} \coloneqq \Scal \times \Hcal$, $\tilde{r}(s\circ h, a) \coloneqq r(s, a)$, $\tilde{\mathbb{P}}(s \circ h, a, s' \circ h') \coloneqq \mathbb{P}(s, a, s') \mbb1{\{h'= \left[ h+1 \modulo H+1\right] \}}$, where $\mbb1{\{\cdot\}}$ is indicator function and $\circ$ denotes concatenation.
\end{definition}

\begin{lemma}\label{lemma:preserve-base}
While applying Algorithm~\ref{alg:bold} as \local, $\tilde{\pi}$ is slowly changing in $\tilde{\mathcal{M}}$, where $\tilde{\pi}_t(a|s \circ h) \coloneqq \pi_t(a|s)$ and $\pi_t$ is produced by \name.
\end{lemma}


Proof of Lemma~\ref{lemma:preserve-base} is deferred to Appendix~\ref{appendix:slowly-changing}.
The switching mechanism in Algorithm~\ref{alg:bold} is now part of the transition function and it is then possible to preserve the slowly changing property. 
The costs are that (1) we increased the cardinality of state space to $\Ocal(HS)$, (2) the stationary distribution is now bounded below by $\Ocal(\beta/H)$, as $\beta$ was a uniform bound which is therefore independent of $t$ and $h$.
For completeness, Lemma~\ref{lemma:stationary-and-mixing} shows that $\tilde{\mathcal{M}}$ satisfies our assumptions on $\mathcal{M}$.
The proof is deferred to Appendix~\ref{appendix:argumented-mdp}. Besides, as $\pi$ uniquely determines $\tilde{\pi}$, we therefore simply use $\pi$ to denote $\tilde{\pi}$ for brevity, for example we write $\tilde{\mu}^\pi$, $\tilde{\Pbb}^\pi$ instead of $\tilde{\mu}^{\tilde{\pi}}$, $\tilde{\Pbb}^{\tilde{\pi}}$.

\begin{lemma}\label{lemma:stationary-and-mixing}
    If assumption~\ref{asm:uniform-bound} and assumption~\ref{asm:fast-mixing} hold for an MDP $\mathcal{M}$, then for its $\Hcal$-augmented counterpart $\tilde{\mathcal{M}}$,
    
    (1) there is an unique stationary distribution $\tilde{\mu}^{{\pi}}$ for any $\pi$;
    
    (2) $\inf_{\pi, \tilde{s}}\tilde{\mu}^\pi(\tilde{s}) \geq \beta/(H+1)$, where $\tilde{s}\in \tilde{\Scal}$; 
    
    (3) $\sup_\pi\|(\tilde{d}-\tilde{d}')\tilde{\Pbb}^{{\pi}}\|_1 \leq e^{-1/\tau}\|\tilde{d}-\tilde{d}'\|_1 $, for any $\tilde{d}$, $\tilde{d}'$.
\end{lemma}

{
\subsection{Main Theorem}
We are now ready to present our main theorem. Theorem~\ref{thm:main} concludes our reduction from RL to adversary bandits, by combining our prior results.

\begin{theorem}\label{thm:main}
    When assumption~\ref{asm:uniform-bound} and assumption~\ref{asm:fast-mixing} hold, apply Algorithm~\ref{alg:bold} as \local, suppose \base\ of Algorithm~\ref{alg:bold} enjoys $\tilde{\Ocal}(f(A)\sqrt{T})$ expect regret in standard adversary setting and is slowly changing with rate $c_T$, then \name~enjoys an expect regret of
    $$ \Reg(T) = 
    \tilde{\Ocal} \left( \frac{H^{2.5} S f(A)}{(1-\gamma) \beta }\sqrt{T} + 
    \frac{\tau^2 H^4 S(S+A)}{(1-\gamma)^2 \beta^3} c_T T \right). $$
    {
    where $f(A)$ is the dependency on $A$ of running \base\ in a standard adversarial non-delayed setting. 
    }
\end{theorem}
\begin{proof}
The regret analysis is structured into aforementioned components. Let's first consider the full-span regret that accumulates all local regrets $R^{\text{fs-mdp}} = \sum_{s\in\Scal} \mathfrak{R}_s(T)$.
\begin{enumerate}

\item Delayed Feedback: Given Corollary~\ref{coroll:delayed-bandits}, {$\local_s$} has a regret of $\tilde{\Ocal}(f(A) \sqrt{HT})$ for state $s$.

\item Sticky Bandits: Since the observed regret at state $s$ is now at most $\tilde{\Ocal}(f(A) \sqrt{HT})$, applying $g(A, H) = f(A)\sqrt{H}$ for Corollary~\ref{coroll:sticky-bandits} leads to 
$ \tilde{\Ocal} \left( \frac{\sqrt{H} S f(A)}{\beta}\sqrt{T} + \left( \frac{S+HA}{(1-\gamma)\beta^3} + \frac{\tau^2}{1-\gamma} \right) c_T T \right). $

\item Objective Mismatch: Corollary~\ref{coroll:obj-mismatch} establishes that the additional error from objective mismatch is at most $\sum_s 2(mc_T T+ \sqrt{T})$, where $m = {H(S+HA)}/{(1-\gamma)}$. It in turn leads to a bound of
$ \tilde{\Ocal} \left( \frac{\sqrt{H} S f(A)}{\beta}\sqrt{T} + \left( \frac{S+HA}{(1-\gamma)\beta^3}  + \frac{\tau^2}{1-\gamma} + \frac{HS(S+HA)}{1-\gamma} \right) c_T T \right).$

\item $\Hcal$-augmented MDPs: As we expand $\Scal$ to $\tilde{\Scal}$, the cardinality of state space is then $\Ocal(HS)$. To accommodate this expansion, we replace $S$ with $HS$ and $\beta$ with $\beta/H$, leading to a bound of 
$ \tilde{\Ocal} \left( \frac{ H^{2.5} S f(A)}{\beta}\sqrt{T} + \left( \frac{H^4(S+A)}{(1-\gamma)\beta^3}  + \frac{\tau^2}{1-\gamma} + \frac{H^3 S(S+A)}{1-\gamma} \right) c_T T \right).$

\end{enumerate}
\noindent We now translate $R^{\text{fs-mdp}}$ into $\Reg(T)$.
\begin{enumerate}
\setcounter{enumi}{4} 
\item Lemma~\ref{lemma:decomposition} establishes that $\Reg(T) \leq \frac{1}{1-\gamma} \sum_{s\in\Scal} \mathfrak{R}_s(T) = \frac{1}{1-\gamma} R^{\text{fs-mdp}}$, leading to the regret bound of 
$ \tilde{\Ocal} \left( \frac{H^{2.5} S f(A)}{(1-\gamma) \beta }\sqrt{T} + 
\frac{\tau^2 H^4 S(S+A)}{(1-\gamma)^2 \beta^3} c_T T
\right).$
%
\end{enumerate}
This concludes our main result.
\end{proof}}
{
\begin{corollary}\label{coroll:main-result-gamma-sim-1}
Suppose the conditions in Theorem~\ref{thm:main} are met. Given that $ H = \Ocal( {\log\sqrt{T}} / {\log(1/\gamma)} ) $, when $ \gamma $ is close to $1$, the bound presented in Theorem~\ref{thm:main} becomes
$ \tilde{\Ocal} \left( \frac{ S f(A)}{(1-\gamma)^{3.5} \beta }\sqrt{T} + 
\frac{\tau^2 S(S+A)}{(1-\gamma)^6 \beta^3} c_T T \right).$
\end{corollary}
}

\section{Case Study: \expt}\label{sec:exp3}
We further extend our discussion on our reduction by providing an example with a well-known exponential-weight bandit algorithm
\expt~\citep{auer2002nonstochastic}.
\subsection{\expt\ as \base}
We first present a regret bound while applying \expt\ as \base\ in our reduction. 
In a standard adversarial non-delayed setting \expt\ has $\tilde{\Ocal}(\sqrt{AT})$ regret~\citep{auer2002nonstochastic} and a slowly-changing rate of $c_T = \tilde{\Ocal}(\sqrt{1/AT})$. 
Applying Theorem~\ref{thm:main} with aforementioned regret and changing rate leads to Corollary~\ref{coroll:main-exp3}, and discussion on this rate $c_T$ can be found in Section~\ref{sec:exp3-rate} and Appendix~\ref{apx:exp3}.
{
\begin{corollary}\label{coroll:main-exp3}
    Applying \expt\ as \base, \name\ has a regret bound of 
     $\tilde{\Ocal} \left( \frac{\tau^2 H^4 S(S+A)}{(1-\gamma)^2\beta^3} \sqrt{T} \right)$,
     which becomes $\tilde{\Ocal} \left( \frac{\tau^2 S(S+A)}{(1-\gamma)^6\beta^3} \sqrt{T} \right)$, when $\gamma$ is close to $1$.
\end{corollary}}
\vspace{-0.1in}
\subsection{\expt\ as \local}\label{sec:exp3-delay}
It is known that the optimal regret is $\tilde{\Ocal} ( \sqrt{(A+z)T} )$ for constant delay $z$~\citep{cesa2016delay}, and remarkably, \expt~achieves the optimal bound~\citep{thune2019nonstochastic}. Furthermore, for unrestricted delays, \cite{bistritz2019online} and \cite{thune2019nonstochastic} show that \expt~enjoys $ \tilde{\Ocal}(\sqrt{AT+Z})$ 
, where $Z$ is the total delay. 
\expt\ therefore enjoys $\tilde{\Ocal} ( \sqrt{(A+H)T })$ regret under our delay $H$.

In previous sections, we use  Algorithm~\ref{alg:bold} as \local, for the purpose of black-box reduction. 
However, \local\ can be any delay-robust adversarial bandit algorithm, such as \expt.
{ Corollary~\ref{coroll:local-exp3} establishes the result when one use \expt~as \local, refining the dependency on $H$ compared to Corollary~\ref{coroll:main-exp3}, which handles delays in a black-box fashion using Algorithm~\ref{alg:bold}.

\begin{corollary}\label{coroll:local-exp3}
    When using \expt\ as \local, the observed regret of \local\ is $\tilde{\Ocal}(\sqrt{(A+H)T})$, and \name\ meets the slowly changing condition without needing the $\mathcal{H}$-augmented trick, leading to a regret of $\tilde{\Ocal} \left( \frac{\tau^2 (HS^2 + H^2 SA)}{(1-\gamma)^2\beta^3} \sqrt{T} \right)$, which turns to $\tilde{\Ocal} \left( \frac{\tau^2 (S^2 + SA/(1-\gamma))}{(1-\gamma)^3\beta^3} \sqrt{T} \right)$ when $\gamma$ is close to $1$.

\end{corollary}
}
\vspace{-0.1in}

\subsection{\expt~is Slowly Changing}\label{sec:exp3-rate}
It can be shown that \expt~meets the slowly changing requirement with a rate of $\eta_T/A$, where $\eta_T$ is the learning rate of \expt. See Appendix~\ref{apx:exp3} for a pseudocode of \expt\ and the proof of Lemma~\ref{lemma:exp3}.
\begin{lemma}\label{lemma:exp3}
Let $\eta_t$ be the learning rate of \expt, \expt~is slowly changing with a rate of $\Ocal(\eta_T/A)$, assuming the feedback is bounded within the range $[0, {1}/{(1-\gamma)}]$.
\end{lemma}

We note that to achieve $\tilde{\Ocal}(\sqrt{AT})$ regret \expt\ is run with a learning rate of {$\eta_T = 
\tilde{\Ocal}{(\sqrt{A/T})}$}, which means it is slowly changing with a rate of $c_T = \tilde{\Ocal}(\sqrt{1/AT})$. 

\section{Conclusion}
{

In this work, we explore the mathematical connections between RL and bandits, in a natural decentralized setting. Our result could serve as a theoretical tool to facilitate generalizing existing bandit results to MDPs, as demonstrated with the example of delayed feedback in Section~\ref{sec:regret}. It can also be linked to multi-agent RL and Monte Carlo methods, as discussed in Section~\ref{sec:intro} and \ref{sec:related}.
However, our results require additional assumptions, and the parameter dependencies, such as 
 those on $S$ and $H$, could still be improved. We further extend our discussion on these limitations and future directions.

One limitation of our work is the need for two extra assumptions not typically needed for discounted infinite-horizon MDPs. These assumptions ensure that all states are visited sufficiently often, hence making the exploration in MDPs less difficult.  
Yet it remains unclear to us whether more aggressive local exploration or algorithm-dependent exploration incentives for local bandit learners could mitigate the need for these assumptions.
However, from the perspective of Monte Carlo learning, our assumptions play a role akin to the exploring starts in the MCES algorithm, as both ensure adequate exploration. Hence, eliminating such assumptions could be an important direction for our framework with Monte Carlo evaluation. 
Another limitation of our result is the relatively large dependency on parameters such as the effective horizon $H$ and the state space size $S$. In Section~\ref{sec:exp3}, we show that the dependency on $H$ can be refined if one directly applies \expt\ as \local. 
As for the dependency on $S$, we believe that, in our current framework, one could not do better than linear dependency on $S$, as accounting for possible policy changes on all states during the effective horizon $H$ unavoidably creates an additional $S$. 
While our work considers adversarial bandits, the varying feedback is in fact not caused by the environment but by the policy changes of its co-learners. Hence, alleviating the requirement of adversarial bandits to stochastic ones could be another important direction. For example, one may consider stochastic bandit algorithms with pre-defined policy-change times to address the non-stationary feedback, since it is known that stochastic bandits attain optimal bounds when the number of changes is known in advance~\citep{auer2019adaptively}.
Moreover, while we focus on the tabular setting, prior work~\citep{brown2019deep} has shown how algorithms using a regret minimizer in every state, such as CFR~\citep{zinkevich2007regret}, have practical implementations via function approximation, which could be another intriguing direction.

}

%

\acks{ { We thank the reviewers and the meta-reviewer for assessing our paper and for their constructive feedback. This work is supported by the National Science Foundation (NSF) grant CCF-1934915 and the NSF grant ECCS-2217023. Zishun is supported in part by the National Institutes of Health (NIH) grant R01CA258827.} }

\bibliography{reference}

\begin{thebibliography}{79}
\providecommand{\natexlab}[1]{#1}
\providecommand{\url}[1]{\texttt{#1}}
\expandafter\ifx\csname urlstyle\endcsname\relax
  \providecommand{\doi}[1]{doi: #1}\else
  \providecommand{\doi}{doi: \begingroup \urlstyle{rm}\Url}\fi

\bibitem[Agrawal and Jia(2017)]{agrawal2017optimistic}
Shipra Agrawal and Randy Jia.
\newblock Optimistic posterior sampling for reinforcement learning: worst-case regret bounds.
\newblock \emph{Advances in Neural Information Processing Systems}, 30, 2017.

\bibitem[Auer et~al.(1995)Auer, Cesa-Bianchi, Freund, and Schapire]{auer1995gambling}
Peter Auer, Nicolo Cesa-Bianchi, Yoav Freund, and Robert~E Schapire.
\newblock Gambling in a rigged casino: The adversarial multi-armed bandit problem.
\newblock In \emph{Proceedings of IEEE 36th annual foundations of computer science}, pages 322--331. IEEE, 1995.

\bibitem[Auer et~al.(2002)Auer, Cesa-Bianchi, Freund, and Schapire]{auer2002nonstochastic}
Peter Auer, Nicolo Cesa-Bianchi, Yoav Freund, and Robert~E Schapire.
\newblock The nonstochastic multiarmed bandit problem.
\newblock \emph{SIAM journal on computing}, 32\penalty0 (1):\penalty0 48--77, 2002.

\bibitem[Auer et~al.(2008)Auer, Jaksch, and Ortner]{auer2008near}
Peter Auer, Thomas Jaksch, and Ronald Ortner.
\newblock Near-optimal regret bounds for reinforcement learning.
\newblock \emph{Advances in neural information processing systems}, 21, 2008.

\bibitem[Auer et~al.(2019)Auer, Gajane, and Ortner]{auer2019adaptively}
Peter Auer, Pratik Gajane, and Ronald Ortner.
\newblock Adaptively tracking the best bandit arm with an unknown number of distribution changes.
\newblock In \emph{Conference on Learning Theory}, pages 138--158. PMLR, 2019.

\bibitem[Bartlett and Tewari(2009)]{bartlett2012regal}
Peter~L Bartlett and Ambuj Tewari.
\newblock Regal: a regularization based algorithm for reinforcement learning in weakly communicating mdps.
\newblock In \emph{Proceedings of the Twenty-Fifth Conference on Uncertainty in Artificial Intelligence}, pages 35--42, 2009.

\bibitem[Bistritz et~al.(2019)Bistritz, Zhou, Chen, Bambos, and Blanchet]{bistritz2019online}
Ilai Bistritz, Zhengyuan Zhou, Xi~Chen, Nicholas Bambos, and Jose Blanchet.
\newblock Online exp3 learning in adversarial bandits with delayed feedback.
\newblock \emph{Advances in neural information processing systems}, 32, 2019.

\bibitem[Brown et~al.(2019)Brown, Lerer, Gross, and Sandholm]{brown2019deep}
Noam Brown, Adam Lerer, Sam Gross, and Tuomas Sandholm.
\newblock Deep counterfactual regret minimization.
\newblock In \emph{International conference on machine learning}, pages 793--802. PMLR, 2019.

\bibitem[Bubeck et~al.(2013)Bubeck, Perchet, and Rigollet]{bubeck2013bounded}
S{\'e}bastien Bubeck, Vianney Perchet, and Philippe Rigollet.
\newblock Bounded regret in stochastic multi-armed bandits.
\newblock In \emph{Conference on Learning Theory}, pages 122--134. PMLR, 2013.

\bibitem[Cai et~al.(2020)Cai, Yang, Jin, and Wang]{cai2020provably}
Qi~Cai, Zhuoran Yang, Chi Jin, and Zhaoran Wang.
\newblock Provably efficient exploration in policy optimization.
\newblock In \emph{International Conference on Machine Learning}, pages 1283--1294. PMLR, 2020.

\bibitem[Cesa-Bianchi et~al.(2016)Cesa-Bianchi, Gentile, Mansour, and Minora]{cesa2016delay}
Nicol‘o Cesa-Bianchi, Claudio Gentile, Yishay Mansour, and Alberto Minora.
\newblock Delay and cooperation in nonstochastic bandits.
\newblock In \emph{Conference on Learning Theory}, pages 605--622. PMLR, 2016.

\bibitem[Cheng et~al.(2020{\natexlab{a}})Cheng, Combes, Boots, and Gordon]{cheng2020reduction}
Ching-An Cheng, Remi~Tachet Combes, Byron Boots, and Geoff Gordon.
\newblock A reduction from reinforcement learning to no-regret online learning.
\newblock In \emph{International Conference on Artificial Intelligence and Statistics}, pages 3514--3524. PMLR, 2020{\natexlab{a}}.

\bibitem[Cheng et~al.(2020{\natexlab{b}})Cheng, Lee, Goldberg, and Boots]{cheng2020online}
Ching-An Cheng, Jonathan Lee, Ken Goldberg, and Byron Boots.
\newblock Online learning with continuous variations: Dynamic regret and reductions.
\newblock In \emph{International Conference on Artificial Intelligence and Statistics}, pages 2218--2228. PMLR, 2020{\natexlab{b}}.

\bibitem[Cravic et~al.(2023)Cravic, Gast, and Gaujal]{cravic2023decentralized}
Romain Cravic, Nicolas Gast, and Bruno Gaujal.
\newblock Decentralized model-free reinforcement learning in stochastic games with average-reward objective.
\newblock In \emph{Proceedings of the 2023 International Conference on Autonomous Agents and Multiagent Systems}, pages 1230--1238, 2023.

\bibitem[Dann and Brunskill(2015)]{dann2015sample}
Christoph Dann and Emma Brunskill.
\newblock Sample complexity of episodic fixed-horizon reinforcement learning.
\newblock \emph{Advances in Neural Information Processing Systems}, 28, 2015.

\bibitem[Dann et~al.(2017)Dann, Lattimore, and Brunskill]{dann2017unifying}
Christoph Dann, Tor Lattimore, and Emma Brunskill.
\newblock Unifying pac and regret: Uniform pac bounds for episodic reinforcement learning.
\newblock \emph{Advances in Neural Information Processing Systems}, 30, 2017.

\bibitem[Dann et~al.(2019)Dann, Li, Wei, and Brunskill]{dann2019policy}
Christoph Dann, Lihong Li, Wei Wei, and Emma Brunskill.
\newblock Policy certificates: Towards accountable reinforcement learning.
\newblock In \emph{International Conference on Machine Learning}, pages 1507--1516. PMLR, 2019.

\bibitem[Dewanto et~al.(2020)Dewanto, Dunn, Eshragh, Gallagher, and Roosta]{dewanto2020average}
Vektor Dewanto, George Dunn, Ali Eshragh, Marcus Gallagher, and Fred Roosta.
\newblock Average-reward model-free reinforcement learning: a systematic review and literature mapping.
\newblock \emph{arXiv preprint arXiv:2010.08920}, 2020.

\bibitem[Dong et~al.(2022)Dong, Wang, and Ross]{dong2022convergence}
Zixuan Dong, Che Wang, and Keith Ross.
\newblock On the convergence of monte carlo ucb for random-length episodic mdps.
\newblock \emph{arXiv preprint arXiv:2209.02864}, 2022.

\bibitem[Etesami(2022)]{etesami2022learning}
S~Rasoul Etesami.
\newblock Learning stationary nash equilibrium policies in $ n $-player stochastic games with independent chains via dual mirror descent.
\newblock \emph{arXiv preprint arXiv:2201.12224}, 2022.

\bibitem[Even-Dar et~al.(2009)Even-Dar, Kakade, and Mansour]{even2009online}
Eyal Even-Dar, Sham~M Kakade, and Yishay Mansour.
\newblock Online markov decision processes.
\newblock \emph{Mathematics of Operations Research}, 34\penalty0 (3):\penalty0 726--736, 2009.

\bibitem[Fruit et~al.(2018{\natexlab{a}})Fruit, Pirotta, and Lazaric]{fruit2018near}
Ronan Fruit, Matteo Pirotta, and Alessandro Lazaric.
\newblock Near optimal exploration-exploitation in non-communicating markov decision processes.
\newblock \emph{Advances in Neural Information Processing Systems}, 31, 2018{\natexlab{a}}.

\bibitem[Fruit et~al.(2018{\natexlab{b}})Fruit, Pirotta, Lazaric, and Ortner]{fruit2018efficient}
Ronan Fruit, Matteo Pirotta, Alessandro Lazaric, and Ronald Ortner.
\newblock Efficient bias-span-constrained exploration-exploitation in reinforcement learning.
\newblock In \emph{International Conference on Machine Learning}, pages 1578--1586. PMLR, 2018{\natexlab{b}}.

\bibitem[Gerchinovitz and Lattimore(2016)]{gerchinovitz2016refined}
S{\'e}bastien Gerchinovitz and Tor Lattimore.
\newblock Refined lower bounds for adversarial bandits.
\newblock \emph{Advances in Neural Information Processing Systems}, 29, 2016.

\bibitem[He et~al.(2021)He, Zhou, and Gu]{he2020nearly}
Jiafan He, Dongruo Zhou, and Quanquan Gu.
\newblock Nearly minimax optimal reinforcement learning for discounted mdps.
\newblock \emph{Advances in Neural Information Processing Systems}, 34:\penalty0 22288--22300, 2021.

\bibitem[Howson et~al.(2021)Howson, Pike-Burke, and Filippi]{howson2021delayed}
Benjamin Howson, Ciara Pike-Burke, and Sarah Filippi.
\newblock Delayed feedback in episodic reinforcement learning.
\newblock \emph{arXiv preprint arXiv:2111.07615}, 2021.

\bibitem[Jin et~al.(2018)Jin, Allen-Zhu, Bubeck, and Jordan]{jin2018q}
Chi Jin, Zeyuan Allen-Zhu, Sebastien Bubeck, and Michael~I Jordan.
\newblock Is q-learning provably efficient?
\newblock \emph{Advances in neural information processing systems}, 31, 2018.

\bibitem[Jin et~al.(2020)Jin, Jin, Luo, Sra, and Yu]{jin2020learning}
Chi Jin, Tiancheng Jin, Haipeng Luo, Suvrit Sra, and Tiancheng Yu.
\newblock Learning adversarial markov decision processes with bandit feedback and unknown transition.
\newblock In \emph{International Conference on Machine Learning}, pages 4860--4869. PMLR, 2020.

\bibitem[Jin et~al.(2022{\natexlab{a}})Jin, Liu, Wang, and Yu]{jin2022v}
Chi Jin, Qinghua Liu, Yuanhao Wang, and Tiancheng Yu.
\newblock V-learning--a simple, efficient, decentralized algorithm for multiagent rl.
\newblock In \emph{ICLR 2022 Workshop on Gamification and Multiagent Solutions}, 2022{\natexlab{a}}.

\bibitem[Jin et~al.(2022{\natexlab{b}})Jin, Lancewicki, Luo, Mansour, and Rosenberg]{jin2022near}
Tiancheng Jin, Tal Lancewicki, Haipeng Luo, Yishay Mansour, and Aviv Rosenberg.
\newblock Near-optimal regret for adversarial mdp with delayed bandit feedback.
\newblock \emph{Advances in Neural Information Processing Systems}, 35:\penalty0 33469--33481, 2022{\natexlab{b}}.

\bibitem[Joulani et~al.(2013)Joulani, Gyorgy, and Szepesv{\'a}ri]{joulani2013online}
Pooria Joulani, Andras Gyorgy, and Csaba Szepesv{\'a}ri.
\newblock Online learning under delayed feedback.
\newblock In \emph{International Conference on Machine Learning}, pages 1453--1461. PMLR, 2013.

\bibitem[Kakade and Langford(2002)]{kakade2002approximately}
Sham Kakade and John Langford.
\newblock Approximately optimal approximate reinforcement learning.
\newblock In \emph{Proceedings of the Nineteenth International Conference on Machine Learning}, pages 267--274, 2002.

\bibitem[Kakade(2003)]{kakade2003sample}
Sham~Machandranath Kakade.
\newblock \emph{On the sample complexity of reinforcement learning}.
\newblock University of London, University College London (United Kingdom), 2003.

\bibitem[Kearns and Singh(2002)]{kearns2002near}
Michael Kearns and Satinder Singh.
\newblock Near-optimal reinforcement learning in polynomial time.
\newblock \emph{Machine learning}, 49\penalty0 (2):\penalty0 209--232, 2002.

\bibitem[Kolter and Ng(2009)]{kolter2009near}
J~Zico Kolter and Andrew~Y Ng.
\newblock Near-bayesian exploration in polynomial time.
\newblock In \emph{Proceedings of the 26th annual international conference on machine learning}, pages 513--520, 2009.

\bibitem[Lattimore and Hutter(2012)]{lattimore2012pac}
Tor Lattimore and Marcus Hutter.
\newblock Pac bounds for discounted mdps.
\newblock In \emph{Algorithmic Learning Theory: 23rd International Conference, ALT 2012, Lyon, France, October 29-31, 2012. Proceedings 23}, pages 320--334. Springer, 2012.

\bibitem[Lattimore et~al.(2013)Lattimore, Hutter, and Sunehag]{lattimore2013sample}
Tor Lattimore, Marcus Hutter, and Peter Sunehag.
\newblock The sample-complexity of general reinforcement learning.
\newblock In \emph{International Conference on Machine Learning}, pages 28--36. PMLR, 2013.

\bibitem[Li et~al.(2021)Li, Shi, Chen, Gu, and Chi]{li2021breaking}
Gen Li, Laixi Shi, Yuxin Chen, Yuantao Gu, and Yuejie Chi.
\newblock Breaking the sample complexity barrier to regret-optimal model-free reinforcement learning.
\newblock \emph{Advances in Neural Information Processing Systems}, 34, 2021.

\bibitem[Li and Yang(2023)]{li2023horizon}
Shengshi Li and Lin Yang.
\newblock Horizon-free learning for markov decision processes and games: stochastically bounded rewards and improved bounds.
\newblock In \emph{International Conference on Machine Learning}, pages 20221--20252. PMLR, 2023.

\bibitem[Liu(2021)]{liu2021convergence}
Jun Liu.
\newblock On the convergence of reinforcement learning with monte carlo exploring starts.
\newblock \emph{Automatica}, 129:\penalty0 109693, 2021.

\bibitem[Liu and Su(2020)]{liu2020regret}
Shuang Liu and Hao Su.
\newblock Regret bounds for discounted mdps.
\newblock \emph{arXiv preprint arXiv:2002.05138}, 2020.

\bibitem[Machado et~al.(2018)Machado, Bellemare, Talvitie, Veness, Hausknecht, and Bowling]{machado2018revisiting}
Marlos~C Machado, Marc~G Bellemare, Erik Talvitie, Joel Veness, Matthew Hausknecht, and Michael Bowling.
\newblock Revisiting the arcade learning environment: Evaluation protocols and open problems for general agents.
\newblock \emph{Journal of Artificial Intelligence Research}, 61:\penalty0 523--562, 2018.

\bibitem[M{\'e}nard et~al.(2021)M{\'e}nard, Domingues, Shang, and Valko]{menard2021ucb}
Pierre M{\'e}nard, Omar~Darwiche Domingues, Xuedong Shang, and Michal Valko.
\newblock Ucb momentum q-learning: Correcting the bias without forgetting.
\newblock In \emph{International Conference on Machine Learning}, pages 7609--7618. PMLR, 2021.

\bibitem[Modi et~al.(2020)Modi, Jiang, Tewari, and Singh]{modi2020sample}
Aditya Modi, Nan Jiang, Ambuj Tewari, and Satinder Singh.
\newblock Sample complexity of reinforcement learning using linearly combined model ensembles.
\newblock In \emph{International Conference on Artificial Intelligence and Statistics}, pages 2010--2020. PMLR, 2020.

\bibitem[Mondal and Aggarwal(2023)]{mondal2023reinforcement}
Washim~Uddin Mondal and Vaneet Aggarwal.
\newblock Reinforcement learning with delayed, composite, and partially anonymous reward.
\newblock \emph{arXiv preprint arXiv:2305.02527}, 2023.

\bibitem[Neu and Pike-Burke(2020)]{neu2020unifying}
Gergely Neu and Ciara Pike-Burke.
\newblock A unifying view of optimism in episodic reinforcement learning.
\newblock \emph{Advances in Neural Information Processing Systems}, 33:\penalty0 1392--1403, 2020.

\bibitem[Neu et~al.(2010)Neu, Gy{\"o}rgy, Szepesv{\'a}ri, and Antos]{gergely2010online}
Gergely Neu, Andr{\'a}s Gy{\"o}rgy, Csaba Szepesv{\'a}ri, and Andr{\'a}s Antos.
\newblock Online markov decision processes under bandit feedback.
\newblock In \emph{Proceedings of the Twenty-Fourth Annual Conference on Neural Information Processing Systems}, 2010.

\bibitem[Ortner(2020)]{ortner2020regret}
Ronald Ortner.
\newblock Regret bounds for reinforcement learning via markov chain concentration.
\newblock \emph{Journal of Artificial Intelligence Research}, 67:\penalty0 115--128, 2020.

\bibitem[Osband and Van~Roy(2017)]{osband2017posterior}
Ian Osband and Benjamin Van~Roy.
\newblock Why is posterior sampling better than optimism for reinforcement learning?
\newblock In \emph{International conference on machine learning}, pages 2701--2710. PMLR, 2017.

\bibitem[Osband et~al.(2013)Osband, Russo, and Van~Roy]{osband2013more}
Ian Osband, Daniel Russo, and Benjamin Van~Roy.
\newblock (more) efficient reinforcement learning via posterior sampling.
\newblock \emph{Advances in Neural Information Processing Systems}, 26, 2013.

\bibitem[Ouyang et~al.(2017)Ouyang, Gagrani, Nayyar, and Jain]{ouyang2017learning}
Yi~Ouyang, Mukul Gagrani, Ashutosh Nayyar, and Rahul Jain.
\newblock Learning unknown markov decision processes: A thompson sampling approach.
\newblock \emph{Advances in neural information processing systems}, 30, 2017.

\bibitem[Pacchiano et~al.(2021)Pacchiano, Ball, Parker-Holder, Choromanski, and Roberts]{pacchiano2021towards}
Aldo Pacchiano, Philip Ball, Jack Parker-Holder, Krzysztof Choromanski, and Stephen Roberts.
\newblock Towards tractable optimism in model-based reinforcement learning.
\newblock In \emph{Uncertainty in Artificial Intelligence}, pages 1413--1423. PMLR, 2021.

\bibitem[Pike-Burke et~al.(2018)Pike-Burke, Agrawal, Szepesvari, and Grunewalder]{pike2018bandits}
Ciara Pike-Burke, Shipra Agrawal, Csaba Szepesvari, and Steffen Grunewalder.
\newblock Bandits with delayed, aggregated anonymous feedback.
\newblock In \emph{International Conference on Machine Learning}, pages 4105--4113. PMLR, 2018.

\bibitem[Rosenberg and Mansour(2019)]{rosenberg2019online}
Aviv Rosenberg and Yishay Mansour.
\newblock Online stochastic shortest path with bandit feedback and unknown transition function.
\newblock \emph{Advances in Neural Information Processing Systems}, 32, 2019.

\bibitem[Russo(2019)]{russo2019worst}
Daniel Russo.
\newblock Worst-case regret bounds for exploration via randomized value functions.
\newblock \emph{Advances in Neural Information Processing Systems}, 32, 2019.

\bibitem[Shalev-Shwartz et~al.(2012)]{shalev2012online}
Shai Shalev-Shwartz et~al.
\newblock Online learning and online convex optimization.
\newblock \emph{Foundations and Trends{\textregistered} in Machine Learning}, 4\penalty0 (2):\penalty0 107--194, 2012.

\bibitem[Simchowitz and Jamieson(2019)]{simchowitz2019non}
Max Simchowitz and Kevin~G Jamieson.
\newblock Non-asymptotic gap-dependent regret bounds for tabular mdps.
\newblock \emph{Advances in Neural Information Processing Systems}, 32, 2019.

\bibitem[Strehl and Littman(2008)]{strehl2008analysis}
Alexander~L Strehl and Michael~L Littman.
\newblock An analysis of model-based interval estimation for markov decision processes.
\newblock \emph{Journal of Computer and System Sciences}, 74\penalty0 (8):\penalty0 1309--1331, 2008.

\bibitem[Strehl et~al.(2006)Strehl, Li, Wiewiora, Langford, and Littman]{strehl2006pac}
Alexander~L Strehl, Lihong Li, Eric Wiewiora, John Langford, and Michael~L Littman.
\newblock Pac model-free reinforcement learning.
\newblock In \emph{Proceedings of the 23rd international conference on Machine learning}, pages 881--888, 2006.

\bibitem[Sutton and Barto(2018)]{sutton2018reinforcement}
Richard~S Sutton and Andrew~G Barto.
\newblock \emph{Reinforcement learning: An introduction}.
\newblock MIT press, 2018.

\bibitem[Szita and Szepesv{\'a}ri(2010)]{szita2010model}
Istv{\'a}n Szita and Csaba Szepesv{\'a}ri.
\newblock Model-based reinforcement learning with nearly tight exploration complexity bounds.
\newblock In \emph{ICML}, 2010.

\bibitem[Talebi and Maillard(2018)]{talebi2018variance}
Mohammad~Sadegh Talebi and Odalric-Ambrym Maillard.
\newblock Variance-aware regret bounds for undiscounted reinforcement learning in mdps.
\newblock In \emph{Algorithmic Learning Theory}, pages 770--805. PMLR, 2018.

\bibitem[Thune et~al.(2019)Thune, Cesa-Bianchi, and Seldin]{thune2019nonstochastic}
Tobias~Sommer Thune, Nicol{\`o} Cesa-Bianchi, and Yevgeny Seldin.
\newblock Nonstochastic multiarmed bandits with unrestricted delays.
\newblock \emph{Advances in Neural Information Processing Systems}, 32, 2019.

\bibitem[Tsitsiklis(2002)]{tsitsiklis2002convergence}
John~N Tsitsiklis.
\newblock On the convergence of optimistic policy iteration.
\newblock \emph{Journal of Machine Learning Research}, 3\penalty0 (Jul):\penalty0 59--72, 2002.

\bibitem[Wang et~al.(2021)Wang, Yuan, Shao, and Ross]{wang2021convergence}
Che Wang, Shuhan Yuan, Kai Shao, and Keith~W Ross.
\newblock On the convergence of the monte carlo exploring starts algorithm for reinforcement learning.
\newblock In \emph{International Conference on Learning Representations}, 2021.

\bibitem[Wang et~al.(2020{\natexlab{a}})Wang, Du, Yang, and Kakade]{wang2020long}
Ruosong Wang, Simon~S Du, Lin~F Yang, and Sham~M Kakade.
\newblock Is long horizon reinforcement learning more difficult than short horizon reinforcement learning?
\newblock \emph{arXiv preprint arXiv:2005.00527}, 2020{\natexlab{a}}.

\bibitem[Wang et~al.(2020{\natexlab{b}})Wang, Dong, Chen, and Wang]{dong2019q}
Yuanhao Wang, Kefan Dong, Xiaoyu Chen, and Liwei Wang.
\newblock Q-learning with ucb exploration is sample efficient for infinite-horizon mdp.
\newblock In \emph{International Conference on Learning Representations}, 2020{\natexlab{b}}.

\bibitem[Wei et~al.(2021)Wei, Jahromi, Luo, and Jain]{wei2021learning}
Chen-Yu Wei, Mehdi~Jafarnia Jahromi, Haipeng Luo, and Rahul Jain.
\newblock Learning infinite-horizon average-reward mdps with linear function approximation.
\newblock In \emph{International Conference on Artificial Intelligence and Statistics}, pages 3007--3015. PMLR, 2021.

\bibitem[Winnicki and Srikant(2023)]{winnicki2023convergence}
Anna Winnicki and R~Srikant.
\newblock On the convergence of policy iteration-based reinforcement learning with monte carlo policy evaluation.
\newblock In \emph{International Conference on Artificial Intelligence and Statistics}, pages 9852--9878. PMLR, 2023.

\bibitem[Xu et~al.(2020)Xu, Wang, and Liang]{xu2020improving}
Tengyu Xu, Zhe Wang, and Yingbin Liang.
\newblock Improving sample complexity bounds for (natural) actor-critic algorithms.
\newblock \emph{Advances in Neural Information Processing Systems}, 33:\penalty0 4358--4369, 2020.

\bibitem[Yan et~al.(2023)Yan, Li, Chen, and Fan]{yan2023efficacy}
Yuling Yan, Gen Li, Yuxin Chen, and Jianqing Fan.
\newblock The efficacy of pessimism in asynchronous q-learning.
\newblock \emph{IEEE Transactions on Information Theory}, 2023.

\bibitem[Yang et~al.(2021)Yang, Yang, and Du]{yang2021q}
Kunhe Yang, Lin Yang, and Simon Du.
\newblock Q-learning with logarithmic regret.
\newblock In \emph{International Conference on Artificial Intelligence and Statistics}, pages 1576--1584. PMLR, 2021.

\bibitem[Zhang and Ji(2019)]{zhang2019regret}
Zihan Zhang and Xiangyang Ji.
\newblock Regret minimization for reinforcement learning by evaluating the optimal bias function.
\newblock \emph{Advances in Neural Information Processing Systems}, 32, 2019.

\bibitem[Zhang and Xie(2023)]{zhang2023sharper}
Zihan Zhang and Qiaomin Xie.
\newblock Sharper model-free reinforcement learning for average-reward markov decision processes.
\newblock In \emph{The Thirty Sixth Annual Conference on Learning Theory}, pages 5476--5477. PMLR, 2023.

\bibitem[Zhang et~al.(2020)Zhang, Zhou, and Ji]{zhang2020almost}
Zihan Zhang, Yuan Zhou, and Xiangyang Ji.
\newblock Almost optimal model-free reinforcement learningvia reference-advantage decomposition.
\newblock \emph{Advances in Neural Information Processing Systems}, 33:\penalty0 15198--15207, 2020.

\bibitem[Zhang et~al.(2021)Zhang, Ji, and Du]{zhang2021reinforcement}
Zihan Zhang, Xiangyang Ji, and Simon Du.
\newblock Is reinforcement learning more difficult than bandits? a near-optimal algorithm escaping the curse of horizon.
\newblock In \emph{Conference on Learning Theory}, pages 4528--4531. PMLR, 2021.

\bibitem[Zhang et~al.(2022)Zhang, Ji, and Du]{zhang2022horizon}
Zihan Zhang, Xiangyang Ji, and Simon Du.
\newblock Horizon-free reinforcement learning in polynomial time: the power of stationary policies.
\newblock In \emph{Conference on Learning Theory}, pages 3858--3904. PMLR, 2022.

\bibitem[Zhou et~al.(2021)Zhou, He, and Gu]{zhou2021provably}
Dongruo Zhou, Jiafan He, and Quanquan Gu.
\newblock Provably efficient reinforcement learning for discounted mdps with feature mapping.
\newblock In \emph{International Conference on Machine Learning}, pages 12793--12802. PMLR, 2021.

\bibitem[Zinkevich et~al.(2007)Zinkevich, Johanson, Bowling, and Piccione]{zinkevich2007regret}
Martin Zinkevich, Michael Johanson, Michael Bowling, and Carmelo Piccione.
\newblock Regret minimization in games with incomplete information.
\newblock \emph{Advances in neural information processing systems}, 20, 2007.

\end{thebibliography}


\appendix

%

\appendix

\section{A Hard Instance}\label{appendix:counter-example}

\begin{figure*}[h] %
\centering 
\begin{tikzpicture}
\node[state] (s1) {$s_1$};
\node[state, right of=s1] (s2) {$s_2$};
\node[state, right of=s2] (s3) {$s_3$};
\node[right of=s3] (dots) {$\ldots$};
\node[state, right of=dots] (sn) {$s_N$};
\draw
(s1) edge[bend left, above] node{\scriptsize w.p.1, r=0} (s2)

(s2) edge[bend left, above] node{\scriptsize 1/3, 1} (s1)
(s2) edge[bend left, above] node{\scriptsize 1/3, 0} (s3)
(s2) edge[above] node{\scriptsize 1/3, 1} (s1)

(s3) edge[bend left, above] node{\scriptsize 1/3, 1} (s2)
(s3) edge[bend left, above] node{\scriptsize 1/3, 0} (dots)
(s3) edge[above] node{\scriptsize 1/3, 1} (s2)

(dots) edge[bend left, above] node{\scriptsize 1/3, 1} (s3)
(dots) edge[bend left, above] node{\scriptsize 1/3, 0} (sn)
(dots) edge[above] node{\scriptsize 1/3, 1} (s3)

(sn) edge[bend left, above] node{\scriptsize 1, R} (s1);
\end{tikzpicture}
\caption{A hard instance for bandits, $\xrightarrow{p, r}$ denotes transition probability $p$ and reward $r$.}
\label{fig:mdp_instance}
\end{figure*}
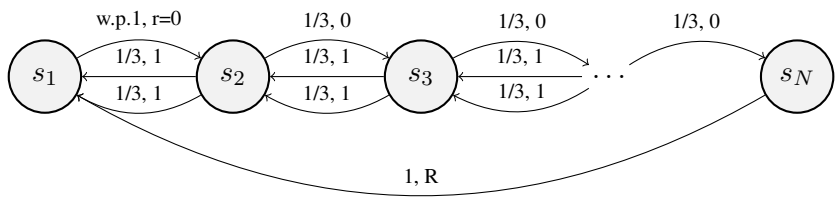

Consider the following deterministic MDP, where most nodes, except $s_1$ and $s_N$, have three actions $l_1, l_2$ and $r$, which stand for action going left and action going right, respectively. And let $S_0 = s_1$ w.p. 1. Going left with $l_1, l_2$ always admits a small reward $1$, but the transition $s_N \rightarrow s_1$ has a large reward $R$. 

Now we place in each state a bandit learner. As bandits are often initialized to assign equal probability to all actions, therefore $p(A_t = r) \leq p(A_t = l_i)$ prior to the first time hitting $s_N$. Therefore, one could consider a uniform policy $\pi(a|s_i) = 1/3$ for $a = l_1, l_2, r$, without losing generality. 

As a result, the Markov chain induced by $\pi$ is equivalent to random walking on positive integers with a biased coin, prior to first hitting $s_N$. Let $M_n$ be the first hitting time of $s_n$, then $\E_\pi[M_n]=\Ocal(2^n)$. Let $N$ be $\Ocal(\sqrt{T})$, as if $N \geq T$ then no policy can be no-regret, (and for example, one could choose $R=\Omega(1/\gamma^T)$ so that the optimal policy is $\pi^*(r|s)=1$). Then, we have $\E_\pi[M_N]=\Ocal(2^{\sqrt{T}}) \geq \Ocal(T)$. The expected first hitting time $\E_\pi[M_N]$ being $\Ocal(2^{\sqrt{T}})$ implies bandits will have $\Ocal(T)$ expected regret. 
Besides, $\Ocal(T)$ first hitting time implies that
this instance is an violation of our assumptions.
Therefore, we need additional assumptions on MDPs  made in Section~\ref{sec:assumptions}. 

But this instance will not be an issue for temperal difference approaches with UCB exploration, for example $\infty$-UCB~\citep{dong2019q}. UCB exploration assigns an exploration bonus to all $(s, a)$ pairs based on the number of visitations of $(s, a)$. %
Therefore, the states on the right will carry a larger bonus because they are rarely visited and the bonus will be propagated via 
temperal difference backups%
to states on the left. As a result, $\infty$-UCB will be encouraged to choose $r$ for exploration, although one has to fine-tune the value of bonus. 
In contrast to our approach, one could consider UCB exploration is centralized as there is a central controller to compute exploration bonus for all state-action pairs $(s, a)$. As our reduction is in a decentralized setting, explorations purely rely on independent bandit learners, which leads to this additional difficulty. 


\section{Technical Tools}\label{sec:tools} 
We first introduce some technical tools, which are useful for our omitted proofs

\begin{lemma}
    $||\cdot||_{1, \infty}$ is a norm
\end{lemma}
\textbf{Proof.} 
Let $X$ and $Y$ be $n$ by $m$ matrices, and $X_{ij}$ be the element corresponding to row $i$ and col $j$

Triangle inequality:
\begin{align}
    ||X+Y||_{1, \infty} &= \max_i \sum_j |(X+Y)_{ij}| \leq \max_i \sum_j (|X_{ij}| + |Y_{ij}|) \\
    &\leq \max_i \sum_j |X_{ij}| + \max_i \sum_j |Y_{ij}| = ||X||_{1, \infty} + ||Y||_{1, \infty}
\end{align}

Absolute homogeneity:
\begin{align}
    ||aX||_{1, \infty} &= \max_i \sum_j |(aX)_{ij}| = |a| \max_i \sum_j |X_{ij}| = |a| \times ||X||_{1, \infty}
\end{align}

Positive definiteness ($||X||_{1, \infty} = 0 \iff X = \mathbf{0}$):
\begin{enumerate}
    \item Let $||X||_{1, \infty} = 0$,
    \begin{align} 
        ||X||_{1, \infty} &= \max_i \sum_j |X_{ij}| = 0
    \end{align}
    implies $X_{ij}=0$ for all $i, j$
    \item Let $X = \mathbf{0}$
    \begin{align} 
        ||\mathbf{0}||_{1, \infty} &= \max_i \sum_j |0| = 0
    \end{align}
\end{enumerate}
Non-negativity:
\begin{align}
    ||X||_{1, \infty} &= \max_i \sum_j |X_{ij}| \geq 0
\end{align}
\qed

One can easily extend the slowly changing property in Definition~\ref{def:slow} to a multi-step version, 
\begin{lemma}\label{lemma:multi-step-slow}
If an algorithm $\A$ is slowly changing with a non-increasing rate of $c_T$, then
    \begin{equation}
        ||\pi_{t+k} - \pi_t||_{1, \infty} \leq k c_T
    \end{equation}
\end{lemma}
\textbf{Proof.} Trivially by triangle inequality.
\begin{align}
    ||\pi_{t+k} - \pi_t||_{1, \infty} &\leq \sum_{i=0}^{k-1} ||\pi_{t+i+1} - \pi_{t+i}||_{1, \infty} 
    \leq \sum_{i=0}^{k-1}c_T \leq kc_T
\end{align}\qed

It is useful to quantify the state distribution gap by following different policies, starting from the same initial state distribution. 
\begin{lemma}\label{lemma:1-step-transit}
    Suppose $||\pi - \pi'||_{1, \infty} \leq c$. Then, for any state distribution vector $d$, we have
    \begin{align}
        ||d\Pbb^\pi - d\Pbb^{\pi'}||_1 &\leq  c
    \end{align}
    where $\Pbb^\pi$ is the transition matrix induced from $\pi$.
\end{lemma}
\textbf{Proof.}
\begin{align}
    ||d\Pbb^\pi - d\Pbb^{\pi'}||_1 &= \sum_{s'} |d\Pbb^\pi(s') - d\Pbb^{\pi'}(s')| \\
    &= \sum_{s'} |\sum_s [d(s)\Pbb^\pi(s, s') - d(s)\Pbb^{\pi'}(s, s')]| \\
    &\leq \sum_{s'} \sum_s d(s) |\Pbb^\pi(s, s') - \Pbb^{\pi'}(s, s')| \\
    &= \sum_{s'} \sum_s d(s) |\sum_a \Pbb(s, a, s')\pi(a|s) - \Pbb(s, a, s')\pi'(a|s)| \\
    &\leq \sum_{s'} \sum_s \sum_a d(s) \Pbb(s, a, s') |\pi(a|s) - \pi'(a|s)| \\
    &= \sum_s d(s) \sum_a |\pi(a|s) - \pi'(a|s)| \\
    &\leq \sum_s d(s) ||\pi-\pi'||_{1, \infty} = c
\end{align}
\qed

Similarly, it is also helpful to bound the state distribution difference after following the same policy, if starting from different state distribution.
\begin{lemma}\label{lemma:contraction}
    For any state distribution vectors $d$ and $d'$, we have
    \begin{align}
        ||d\Pbb^\pi - d'\Pbb^{\pi}||_1 &\leq  ||d-d'||_1
    \end{align}
    where $\Pbb^\pi$ is the transition matrix induced from $\pi$.
\end{lemma}
\textbf{Proof.}
\begin{align}
    ||d\Pbb^\pi - d'\Pbb^{\pi}||_1 &= \sum_{s'} |d\Pbb^\pi(s') - d'\Pbb^\pi(s')|\\
    &= \sum_{s'}|\sum_s [d(s)\Pbb^\pi(s, s') - d'(s)\Pbb^\pi(s, s')]| \\
    &\leq \sum_{s'}\sum_s \Pbb^\pi(s, s') |d(s) - d'(s)| \\
    &= \sum_s |d(s) - d'(s)| \sum_{s'} \Pbb^\pi(s, s') \\
    &= ||d-d'||_1
\end{align}
\qed

The case when starting from different distribution and following different policies for one step.
\begin{lemma}\label{lemma:1-step-error}
    Given policies $\pi$ and $\pi'$, and state distribution vectors $d$ and $d'$, if $|| \pi - \pi' ||_{1, \infty} \leq c$ and $||d - d'||_1 \leq \delta$, then we have
    \begin{equation}
        ||d\Pbb^\pi - d'\Pbb^{\pi'}||_1 \leq  c + \delta
    \end{equation}
\end{lemma}
\textbf{Proof.}
\begin{align}
    ||d\Pbb^\pi - d'\Pbb^{\pi'}||_1 &= 
    ||d\Pbb^\pi - d\Pbb^{\pi'} + d\Pbb^{\pi'} - d'\Pbb^{\pi'}||_1 \\
    &\leq ||d\Pbb^\pi - d\Pbb^{\pi'}||_1 + ||d\Pbb^{\pi'} - d'\Pbb^{\pi'}||_1 \\
    \intertext{by Lemma~\ref{lemma:contraction} and Lemma~\ref{lemma:1-step-transit}}
    &\leq ||d\Pbb^\pi - d\Pbb^{\pi'}||_1 + ||d - d'||_1 \\
    &\leq c + \delta
\end{align}
\qed 


\section{Key Technical Lemma}
Extension to $n$-step case
\begin{lemma}\label{lemma:error-propagate}
    Given two set of policies (of equal size) \{$\pi_1, \dots, \pi_k, \dots, \pi_K$\} and \{$\pi'_1, \dots, \pi_k', \dots, \pi'_K$\} and initial state distribution vectors $d$ and $d'$. If $|| \pi_k - \pi_k' ||_{1, \infty} \leq c $ and $||d - d'||_1 \leq \delta$, then we have
    \begin{align}
        ||d(\Pbb^{\pi_1}\cdots\Pbb^{\pi_K}) - d'(\Pbb^{\pi'_1} \cdots \Pbb^{\pi'_K})||_1 &\leq  Kc + \delta
    \end{align}
\end{lemma}
\textbf{Proof.} We prove this by induction on $K$,
\begin{align}
    \intertext{$K=1$: by Lemma~\ref{lemma:1-step-error}}
    &||d\Pbb^{\pi_1} - d'\Pbb^{\pi'_1}||_1 \leq  c + \delta 
    \intertext{$K=n$: assume we have,}
    &||d(\Pbb^{\pi_1}\cdots\Pbb^{\pi_{n}}) - d'(\Pbb^{\pi'_1} \cdots \Pbb^{\pi'_{n}})||_1 \leq  nc + \delta  
    \intertext{$K=n+1$:}
    &||d(\Pbb^{\pi_1}\cdots\Pbb^{\pi_{n+1}}) - d'(\Pbb^{\pi'_1} \cdots \Pbb^{\pi'_{n+1}})||_1 \\
    &= ||d_{n} \Pbb^{\pi_{n+1}} - d'_{n} \Pbb^{\pi'_{n+1}}|| \\
    &\leq nc + \delta  + c
\end{align}
\qed

\begin{corollary}\label{coroll:transit-multi-Q}
    Let $\pi_t$ be slowly changing with non-increasing rate $c_T$, then we have $||\pi_{t+n} - \pi_t||_{1, \infty} \leq n c_T$. Apply Lemma~\ref{lemma:error-propagate} with $\pi_k = \pi_{t+n}$, $\pi_k' = \pi_t$ and $d=d'$, then
    \begin{align}
        ||d(\Pbb^{\pi_{t+n}})^K - d(\Pbb^{\pi_t})^K||_1  \leq Kn c_T
    \end{align}
\end{corollary}

\begin{corollary}\label{coroll:transit-multi-v1}
    Let $\pi_t$ be slowly changing with non-increasing rate $c_T$, then we have $||\pi_{t+n} - \pi_t||_{1, \infty} \leq n c_T$. Apply Lemma~\ref{lemma:error-propagate} with $\pi_k = \pi_{t+k-1}$, $\pi_k' = \pi_t$ and $d=d'$, then
    \begin{align}
        ||d\underbrace{(\Pbb^{\pi_t}\cdots\Pbb^{\pi_{t+K-1}})}_{\text{$K$ transition kernels}} - d(\Pbb^{\pi_t})^K||_1  \leq K^2 c_T
    \end{align}
\end{corollary}

\begin{corollary}\label{coroll:transit-multi-v2}
    Let $\pi_t$ be slowly changing with non-increasing rate $c_T$, then we have $||\pi_{t+n} - \pi_{t}||_{1, \infty} \leq n c_T$. Apply Lemma~\ref{lemma:error-propagate} with $\pi_k = \pi_{t+n+k-1}$, $\pi_k' = \pi_{t+k-1}$ and $d=d'$, then
    \begin{align}
        ||d\underbrace{(\Pbb^{\pi_{t+n}}\cdots\Pbb^{\pi_{t+n+K-1}})}_{\text{$K$ kernels}} - d\underbrace{(\Pbb^{\pi_t} \cdots \Pbb^{\pi_{t+K-1}})}_{\text{$K$ kernels}} ||_1  \leq Kn c_T
    \end{align}
\end{corollary}

These lemmas and corollaries describe how the state distribution would change by following different sequence of policies, which will be eventually used to prove Lemma~\ref{lemma:approx-error} that bounds $|\bar{U}_t(s, a) - Q_t(s, a)|$ and Lemma~\ref{lemma:delay-error-Q} that bounds $|Q_{t+n}(s, a) - Q_t(s, a)|$.

\section{Proof of Lemma~\ref{lemma:decomposition} (Decomposition Lemma)}\label{sec:decomposition}

We first introduce Performance Difference Lemma~\citep{kakade2002approximately, kakade2003sample}

\begin{lemma}[Performance Difference Lemma.]\label{lem:pdl}
Let $M$ be an MDP, then for all stationary policies $\pi$ and $\pi'$, and for all $s_0$ and $\gamma$,
\begin{align}\label{lemma:pdl}
    V^{\pi'}(s_0) - V^\pi(s_0) = \tfrac{1}{1-\gamma} \E_{s\sim d^{\pi'}_{s_0}} \E_{a\sim \pi'}[Q^\pi(s, a)-V^\pi(s)] \nonumber
\end{align}
where $d^{\pi'}_{s_0}(s) = (1-\gamma)\sum_{t=0}^\infty\gamma^{t}\Pr(S_t = s|\pi', M, S_0 = s_0)$ is the normalized discounted occupancy measure starting from $s_0$ and following $\pi'$.
\end{lemma}

\begin{replemma}{lemma:decomposition}
    The expected regret in MDPs can be reduced to cumulative local regret with oracle feedback $Q_t$
    \begin{equation}
        \Reg(T) \leq \tfrac{1}{1-\gamma} \sum_{s \in \Scal} \mathfrak{R}_s(T)
    \end{equation}
\end{replemma}
\textbf{Proof.}
Let $\{ \pi_t: 1\leq t \leq T \}$ be the sequence of policies obtained by running by any algorithm $\A$.
\begin{align}
    \Reg(T) &= \sT V^*(s_t) - V_t(s_t) 
    \intertext{apply Performance Difference Lemma~\ref{lem:pdl} with $\pi' = \pi^*$ and $\pi = \pi_t$}
    &= \tfrac{1}{1-\gamma} \sT \E_{s\sim d^{\pi^*}_{s_t}} \E_{a\sim \pi^*}[Q_t(s, a)-V_t(s)] \\
    &= \tfrac{1}{1-\gamma} \sT \E_{s\sim d^{\pi^*}_{s_t}} \sum_{a \in \Acal} (\pi^*(a|s) - \pi_t(a|s) )Q_t(s, a) \\
    &\leq \tfrac{1}{1-\gamma} \sT \sum_{s \in \Scal} \sum_{a \in \Acal} (\pi^*(a|s) - \pi_t(a|s) )Q_t(s, a) \\
    \intertext{by definition~\ref{def:local-regret}}
    &=\tfrac{1}{1-\gamma} \sum_{s \in \Scal} \mathfrak{R}_s(T)
\end{align}
\qed

\section{Proof of Lemma~\ref{lemma:approx-error} (Objective Mismatch)}\label{appendix:obj-mismatch}
As the local regret considers objective of state-value function $Q_t$ but our Monte Carlo estimator approximate the target $\bar{U}_t$, we now show the gap is bounded. 

We define $\bar{Q}_t \coloneqq \E[\sum_{t=0}^H \gamma^t R_t|\pi, S_0 = s, A_0 = a]$, a finite horizon counterpart of $Q_t$, we will use this notation for proof in later sections as well.
We first show the gap between $Q_t$ and $\bar{Q}_t$, as it is easier to compare $\bar{U}_t$ and $\bar{Q}_t$ because of the same finite horizon

\begin{lemma}\label{lemma:general_tail}
    Let $\{X_i, i=1, \cdots, \}$ be an arbitrary infinite sequence such that $X_i \in [0, 1]$ for all $i$, $\gamma$ be the discounted factor of a MDP, we have
    \begin{equation}
        \left| \sum_{t=0}^\infty \gamma^t X_t - \sum_{t=0}^H \gamma^t X_t \right| \leq \frac{1}{\sqrt{T}}
    \end{equation}
\end{lemma}
\textbf{Proof.}
\begin{align}
    \left| \sum_{t=0}^\infty \gamma^t X_t - \sum_{t=0}^H \gamma^t X_t \right| &= \left| \sum_{t=H+1}^\infty \gamma^t X_t \right| \\
    &= \sum_{t=H+1}^\infty \gamma^t = \frac{\gamma^{H+1}}{1-\gamma}
    \intertext{by the definition of $H$ in Algo.~\ref{alg:main}}
    & \leq \frac{\gamma^{\log_\gamma{({1-\gamma})/{\sqrt{T}}}}}{1-\gamma} = \frac{1}{\sqrt{T}}
\end{align}
\qed

\begin{corollary}\label{lemma:tail}
    As a result of Lemma~\ref{lemma:general_tail}, we have
    \begin{align}
        |\bar{Q}_t(s, a) - Q_t(s, a)| \leq \frac{1}{\sqrt{T}}, |\bar{U}_t(s, a) - U_t(s, a)| \leq \frac{1}{\sqrt{T}}.
    \end{align}
\end{corollary}

Before giving the first key lemma, we first note a fact that $|a_1 b_1 - a_2 b_2| \leq |a_1 - a_2| + |b_1 - b_2|$ when $a_1, a_2, b_1, b_2 \in [0, 1]$. Let $a_1, a_2, b_1, b_2 \in [0, 1]$, we have
%
%
\begin{align}
    |a_1 b_1 - a_2 b_2| &=  |a_1 b_1 - a_1 b_2 + a_1 b_2 - a_2 b_2| \\
    &\leq |a_1 b_1 - a_1 b_2| + |a_1 b_2 - a_2 b_2| \\
    &\leq |a_1 - a_2| + |b_1 - b_2|
\end{align}

Now we are ready to give the first key lemma
\begin{replemma}{lemma:approx-error}
    If \name~is slowly changing with a non-increasing rate of $c_T$, then 
    \begin{equation}
        |\bar{U}_t(s, a) - Q_t(s, a)| \leq \frac{H S + H^2 A}{1-\gamma} c_T + \frac{1}{\sqrt{T}}
    \end{equation}
\end{replemma}
\textbf{Proof.}
Let $k=0, 1, 2, \dots$, and $U_t$ as defined in the main text. And recall that Corollary~\ref{coroll:transit-multi-v1} described the state distribution gap after running different sequence of policies, starting from same distribution, shown as below

\begin{center}
\fbox{%
\parbox{0.88\textwidth}{
    \begin{repcorollary}{coroll:transit-multi-v1}
        Let $\pi_t$ be slowly changing with non-increasing rate $c_T$, then we have $||\pi_{t+l} - \pi_t||_{1, \infty} \leq l c_T$. Apply Lemma~\ref{lemma:error-propagate} with $\pi_k = \pi_{t+k-1}$, $\pi_k' = \pi_t$ and $d=d'$, then
        \begin{align}
            ||d\underbrace{(\Pbb^{\pi_t}\cdots\Pbb^{\pi_{t+K-1}})}_{\text{$K$ transition kernels}} - d(\Pbb^{\pi_t})^K||_1  \leq K^2 c_T
        \end{align}
    \end{repcorollary}
}}
\end{center}

Noticing that the following facts perfectly fit the conditions of applying Corollary~\ref{coroll:transit-multi-v1}
\begin{itemize}
    \item $\pi_t$ is used to evaluate $Q_t(s,a)$, $\{\pi_t, \dots, \pi_{t+k}, \dots\}$ are used to evaluate $U_t(s, a)$, and the initial policy is the same $\pi_t$
    \item The initial state are both $s$, because we are evaluating $Q_t(s, a)$ and $U_t(s, a)$. Let this deterministic distribution be $d_s$.
    \item The future state distributions $d_{k}$ (that used to evaluate $U_t$) follows $d_{k+1} = d_{k}\Tcal^{\pi_{t+k}}$
    \item The future state distributions $d'_{k}$ (that used to evaluate $Q_t$) follows $d'_{k+1} = d'_{k}\Tcal^{\pi_{t}}$
\end{itemize}
We introduce the symbols $d_s$ and $d_k$ instead of using $d_t$ and $d_{t+k}$ as to avoid confusion with the actual state distributions produced by running \name. $d_k$ indicates k steps in the future starting from $d_s$, where $s$ is \textbf{not} necessarily the actual visited state at time $t$. Therefore, the notions $d_k$ and $d_s$ are less attached with the distributions realized by algorithm.

Therefore, by Lemma~\ref{lemma:multi-step-slow} and Corollary~\ref{coroll:transit-multi-v1}, we have 
\begin{align}
    & ||\pi_{t+k} - \pi_t||_{1, \infty} \leq k c_T \label{eq:pi_bound_1} \\
    & ||d_s \underbrace{(\Tcal^{\pi_t} \cdots \Tcal^{\pi_{t+k-1}})}_{\text{$k$ kernels}} - d_s (\Tcal^{\pi_t})^k||_1 \leq k^2 c_T \label{eq:d_bound_1}
\end{align}

Now we are ready to bound the difference between objectives, note that we will use the prime notion $d_{k}$ and $d'_{k}$ as used in the bullet-points above. And we slightly abuse notation here by denoting an element of a vector by e.g. $d'_k(x)$, as both subscript and superscript are occupied
\begin{align}
    \Big |\bar{U}_t(s, a) - Q_t(s, a) \Big| &\leq \Big |\bar{U}_t(s, a) - \bar{Q}_t(s, a) \Big| + \frac{1}{\sqrt{T}}\\
    &= \B| \sum_{k=0}^H \sum_{x\in\Scal} \sum_{i\in\Acal} \gamma^k 
    \B[ d_{k}(x) \pi_{t+k}(i|x) - d'_{k}(x) \pi_t(i|x) \B] r(x, i) \B|  + \frac{1}{\sqrt{T}}\\
    &\leq  \sum_{k=0}^H \sum_{x, i} \gamma^k \B|  
    \B[ d_{k}(x) \pi_{t+k}(i|x) - d'_{k}(x) \pi_t(i|x) \B] r(x, i) \B|  + \frac{1}{\sqrt{T}}
    \intertext{by the assumption $r(s, a) \in [0, 1]$}
    &\leq  \sum_{k=0}^H \sum_{x, i} \gamma^k \B|  
     d_{k}(x) \pi_{t+k}(i|x) - d'_{k}(x) \pi_t(i|x) \B|  + \frac{1}{\sqrt{T}}
    \intertext{by the fact $|a_1b_1 - a_2b_2| \leq |a_1 - a_2| + |b_1 - b_2|$ for $a_i, b_i \in [0, 1]$}
    &\leq \sum_{k=0}^H \sum_{x, i} \gamma^k \bigg\{ \B|d_{k}(x) - d'_{k}(x) \B| + \B| \pi_{t+k}(i|x) - \pi_t(i|x) \B| \bigg\}  + \frac{1}{\sqrt{T}}\\
    &\leq \sum_{k=0}^H \gamma^k \bigg\{ \sum_x\sum_i \B|d_{k}(x) - d'_{k}(x) \B| + \sum_x \max_x \sum_i \B| \pi_{t+k}(i|x) - \pi_t(i|x) \B| \bigg\}  + \frac{1}{\sqrt{T}}\\
    &= \sum_{k=0}^H \gamma^k \bigg\{ A || d_{k} - d'_{k} ||_1 + S || \pi_{t+k} - \pi_t ||_{1, \infty} \bigg\}  + \frac{1}{\sqrt{T}}
    \intertext{given $d_{k} = d_s(\Tcal^{\pi_{t}}\cdots\Tcal^{\pi_{t+k-1}})$ and $d'_{k} = d_s (\Tcal^{\pi_t})^{k}$}
    &= \sum_{k=0}^H \gamma^k \bigg\{ A \B|\B| d_s(\Tcal^{\pi_{t}}\cdots\Tcal^{\pi_{t+k-1}}) - d_s (\Tcal^{\pi_t})^{k} \B|\B|_1 + S || \pi_{t+k} - \pi_t ||_{1, \infty} \bigg\}  + \frac{1}{\sqrt{T}}
    \intertext{follow Eq.\eqref{eq:pi_bound_1} and Eq.~\eqref{eq:d_bound_1}}
    &\leq \sum_{k=0}^H \gamma^k (S k c_T + A k^2 c_T)  + \frac{1}{\sqrt{T}}
    \intertext{by $k \leq H$ and the sum of geometric sequence}
    &\leq \frac{H S + H^2 A}{1-\gamma} c_T + \frac{1}{\sqrt{T}}
\end{align}

\qed

\section{Proof of Lemma~\ref{lemma:delay-error-Q}}\label{appendix:delay-error-Q}
\begin{replemma}{lemma:delay-error-Q}
    If \name~is slowly changing with a non-increasing rate of $c_T$, then 
    \begin{equation}
        |Q_{t+n}(s, a) - Q_t(s, a)| \leq \frac{1}{1-\gamma} \left(S+HA\right) n c_T + \frac{2}{\sqrt{T}}
    \end{equation}
\end{replemma}
\textbf{Proof:}
Recall
\begin{center}
\fbox{%
\parbox{0.88\textwidth}{
    \begin{repcorollary}{coroll:transit-multi-Q}
        Let $\pi_t$ be slowly changing with non-increasing rate $c_T$, then we have $||\pi_{t+l} - \pi_t||_{1, \infty} \leq l c_T$. Apply Lemma~\ref{lemma:error-propagate} with $\pi_k = \pi_{t+n}$, $\pi_k' = \pi_t$ and $d=d'$, then
    \begin{align}
        ||d(\Pbb^{\pi_{t+n}})^K - d(\Pbb^{\pi_t})^K||_1  \leq Kn c_T
    \end{align}
    \end{repcorollary}
}}
\end{center}

By Lemma~\ref{lemma:multi-step-slow} and corollary~\ref{coroll:transit-multi-Q}, we have 
\begin{align}
    &||\pi_{t+n} - \pi_{t}||_{1, \infty} \leq nc_T \label{eq:pi_bound_q}\\
    &||d_s(\Tcal^{\pi_{t+n}})^k - d_s(\Tcal^{\pi_{t}})^k ||_1 \leq nk c_T \label{eq:d_bound_q}
\end{align}

\begin{align}
    \B| Q_{t+n}(s, a) - Q_t(s, a) \B| 
    &\leq \B| \bar{Q}_{t+n}(s, a) - \bar{Q}_t(s, a) \B| + \B| \bar{Q}_{t+n}(s, a) - Q_{t+n}(s, a) \B| + \B| \bar{Q}_{t}(s, a) - Q_{t}(s, a) \B|\\
    \intertext{by Corollary~\ref{lemma:tail}, we have}
    &\leq \B| \sum_{k=0}^H \sum_{x\in\Scal} \sum_{i\in\Acal} \gamma^k \B[ d'_k(x)\pi_{t+n}(i|x) - d_k(x)\pi_t(i|x) \B] r(x, i) \B| + \frac{2}{\sqrt{T}}\\
    &\leq \sum_{k=0}^H \sum_{x} \sum_{i} \gamma^k \B|   \B[ d'_k(x)\pi_{t+n}(i|x) - d_k(x)\pi_t(i|x) \B] r(x, i) \B|  + \frac{2}{\sqrt{T}}
    \intertext{$r(x, i) \in [0, 1]$}
    &\leq \sum_{k=0}^H \sum_{x} \sum_{i} \gamma^k \B| d'_k(x)\pi_{t+n}(i|x) - d_k(x)\pi_t(i|x) \B|  + \frac{2}{\sqrt{T}}
    \intertext{by the fact $|a_1b_1 - a_2b_2| \leq |a_1 - a_2| + |b_1 - b_2|$ for $a_i, b_i \in [0, 1]$}
    &\leq \sum_{k=0}^H \sum_{x} \sum_{i} \gamma^k \bigg\{ \B| d'_k(x) - d_k(x) \B| + \B| \pi_{t+n}(i|x) -\pi_t(i|x) \B| \bigg\}  + \frac{2}{\sqrt{T}}\\
    &\leq \sum_{k=0}^H \gamma^k \bigg\{ \sum_{x} \sum_{i} \B| d'_k(x) - d_k(x) \B| + \sum_{x} \max_x \sum_{i} \B| \pi_{t+n}(i|x) -\pi_t(i|x) \B| \bigg\}  + \frac{2}{\sqrt{T}}\\
    &= \sum_{k=0}^H \gamma^k \bigg\{  A||d'_k - d_k||_1 + S ||\pi_{t+n} -\pi_t ||_{1, \infty} \bigg\}  + \frac{2}{\sqrt{T}}\\
    &= \sum_{k=0}^H \gamma^k \bigg\{  A\B|\B|d_s(\Pbb^{\pi_{t+n}})^k - d_s(\Pbb^{\pi_{t}})^k\B|\B|_1 + S ||\pi_{t+n} -\pi_t ||_{1, \infty} \bigg\}  + \frac{2}{\sqrt{T}}\\
    &\leq \sum_{k=0}^H \gamma^k  \B( n S + nk A \B) c_T  + \frac{2}{\sqrt{T}}
    \intertext{by $k\leq H$ and the sum of geometric sequences }
    &\leq \frac{1}{1-\gamma} \left(S+HA\right) n c_T + \frac{2}{\sqrt{T}}
\end{align}

\qed

\section{Proof of Lemma~\ref{lemma:mixing-error}}\label{appendix:mixing-error}
We follow the proof by~\cite{even2009online, gergely2010online}, to give a lemma that $\nu_t$ tracks the stationary distribution $\mu_t$ slowly

\begin{replemma}{lemma:mixing-error}
    If the sequence of policies $\{ \pi_1, \dots \pi_t \}$ is slowly changing with rate $c_T$, then $||\nu_t - \mu_t||_1 \leq \tau(\tau+1)c_T + 2e^{-(t-1)/\tau} $
\end{replemma}
\textbf{Proof.} 
Suppose $k+1 \leq t$
\begin{align}
    ||\nu_{k+1} - \mu_{t}||_1 &= ||\nu_{k}\Pbb^{\pi_k} - \nu_{k}\Pbb^{\pi_t} + \nu_{k}\Pbb^{\pi_t}  - \mu_{t}\Pbb^{\pi_t}||_1 \\
    &\leq ||\nu_{k}\Pbb^{\pi_{k}} - \nu_{k}\Pbb^{\pi_t}||_1 + ||\nu_{k}\Pbb^{\pi_t}  - \mu_{t}\Pbb^{\pi_t}||_1 \\
    \intertext{by Assumption~\ref{asm:fast-mixing}}
    &\leq ||\nu_{k}\Pbb^{\pi_{k}} - \nu_{k}\Pbb^{\pi_t}||_1 + e^{-1/\tau}||\nu_{k} - \mu_t||_1
    \intertext{by Lemma~\ref{lemma:1-step-transit}}
    &\leq (t-k)c_T + e^{-1/\tau}||\nu_{k} - \mu_t||_1
\end{align}
Then, by expanding the recursion we have 
\begin{align}
    ||\nu_t - \mu_{t}||_1 &\leq c_T \sum_{k=1}^{t-1} (t-k)e^{-(t-k-1)/\tau} + e^{-(t-1)/\tau}||\nu_1 - \mu_t ||_1  
    \intertext{by $||\nu_1 - \mu_t||_1 \leq 2$}
    &\leq c_T \sum_{k=1}^{t-1} (t-k)e^{-(t-k-1)/\tau} + 2e^{-(t-1)/\tau}  
    \intertext{notice that $\sum_{k=1}^{t-1} (t-k)e^{-(t-k-1)/\tau} \leq \int_0^\infty (k+1) e^{-k/\tau} dk = \tau^2$}
    &\leq \tau (\tau+1) c_T  + 2e^{-(t-1)/\tau}
\end{align}
\qed

\section{Proof of Lemma~\ref{lemma:full-regret} (Full-Span Regret)}
We first give two technical lemmas, by consider a single sticky bandit

\begin{lemma}\label{lemma:expectation}
    Let $X_i$ be the timestep of $i$-th time that a sticky bandit could react, and suppose each time $t$, the probability of the bandit could react is at least $\beta$, while assuming each draw is independent
    \begin{equation}
        \E_{X_{i+1}|X_i}[X_{i+1}-X_{i}] \leq 1/\beta
    \end{equation}  
\end{lemma}
\textbf{Proof.}
As $X_{i+1} - X_i \geq 1$
\begin{align}
    \E_{X_{i+1}|X_i}(X_{i+1} - X_i) &\leq \beta \times 1 + (1-\beta) \B[ \E_{X_{i+1}|X_i}(X_{i+1} - X_i) + 1 \B] \\
    &\Rightarrow \\
    \E_{X_{i+1}|X_i}(X_{i+1} - X_i) &\leq 1/\beta
\end{align}
\qed

\begin{lemma}\label{lemma:variance}
    Let $X_i$ be the timestep of $i$-th time that a sticky bandit could react, and suppose each time $t$, the probability of the bandit could react is at least $\beta$, while assuming each draw is independent
    \begin{equation}
        \E_{X_{i+1}|X_i} (X_{i+1} - X_{i})^2 \leq \frac{2}{\beta^3}
    \end{equation}  
\end{lemma}
\textbf{Proof.}
\begin{align}
    \E_{X_{i+1}|X_i} (X_{i+1} - X_{i})^2 &= \sum_{t = X_i + 1}^{\infty} \Pr(X_{i+1}=t) (t - X_{i})^2 
    \intertext{as $\Pr(X_{i+1}=t) \geq \beta$ for all $t$}
    &\leq \sum_{t = X_i + 1}^{\infty} \left[ (1-\beta)^{t-X_i-1}\times 1 \right] (t - X_{i})^2
    \intertext{let $k=X_i+1$, $q=1-\beta$ and given $X_i > 0$, we have}
    &= \sum_{t=k}^\infty q^{t-k}(t-k+1)^2 \\
    &= \sum_{t=0}^\infty q^{t}(t+1)^2
    \intertext{let $S_n = \sum_{t=0}^n q^t(t+1)^2$}
    &= \lim_{n\to\infty} S_n \\
    &= \lim_{n\to\infty} \frac{1-q}{1-q} S_n
    \intertext{observe that $S_n - qS_n = 1 - q^{n+1}(n+1)^2 + \sum_{t=1}^n (2t+1)q^t$}
    &= \frac{1}{1-q} \lim_{n\to\infty} \left[ 1 - q^{n+1}(n+1)^2 + \sum_{t=1}^n (2t+1)q^t \right] \\
    &= \frac{1}{1-q} \lim_{n\to\infty} \sum_{t=0}^n (2t+1)q^t 
    \intertext{where $\sum_{t=0}^n (2t+1)q^t$ is an arithmetico-geometric series, by the sequence sum of arithmetico–geometric series}
    &\leq \frac{\beta^2-3\beta +2}{(1-\beta)\beta^3} 
    \leq \frac{2}{\beta^3}
\end{align}
\qed

Now we are ready to bound the full-span regret by the observed regret. 

\begin{replemma}{lemma:full-regret}
    If Assumption~\ref{asm:uniform-bound} and \ref{asm:fast-mixing} are satisfied, and the \local\ learner is slowly changing with a rate $c_T$, we have
    \begin{equation}
        \tilde{R}^{\text{un-mdp}}(T) \leq \frac{\tilde{R}^{\text{ob-mdp}}(T)}{\beta} + \frac{S+HA}{\beta^3(1-\gamma)} c_T T + 4S\sqrt{T} .
    \end{equation}
\end{replemma}
\textbf{Proof.}
Recall that the definitions of observed/unobserved regret in MDPs are,
\begin{align}
    \tilde{R}^{\text{ob-mdp}} &\coloneqq \sum_{s \in \Scal} \sum_{t=1}^T \mu_t \langle \pi^*_s-\pi_{t, s}, Q_{t, s} \rangle \\
    &=  \sum_{s \in \Scal} \E_{\{Y_i^s\}} \sum_{t\in\{Y_i^s\}} \langle \pi^*_s-\pi_{t, s}, Q_{t, s} \rangle \\
    \tilde{R}^{\text{un-mdp}} &\coloneqq \sum_{s \in \Scal} \sum_{t=1}^T (1-\mu_t) \langle \pi^*_s-\pi_{t, s}, Q_{t, s} \rangle \\
    &=  \sum_{s \in \Scal} \E_{\{Y_i^s\}} \sum_{t\notin \{Y_i^s\}} \langle \pi^*_s-\pi_{t, s}, Q_{t, s} \rangle
\end{align}
where $Y_i^s$ is a random variable that stands for the time step $t$ the bandit at $s$ was allowed to pull, 
{ and $\pi_{t, s} \coloneqq \pi_t(\cdot|s)$ and $Q_{t, s}\coloneqq Q_t(s, \cdot)$ denote vectors corresponding to state $s$ for the sake of space.}

One could divide the sequence into segments $(Y^s_i, Y^s_{i+1})$, and without loss of generality, we assume bandit at $s$ is pulled $N_s$ times in total, define $Y^s_0 = 1$
\begin{align}
    \tilde{R}^{\text{un-mdp}} &=  \sum_s \E_{\{Y_i^s\}} \sum_{i=0}^{N_s} \sum_{t \in (Y^s_i, Y^s_{i+1})} \B[\langle \pi^*_s-\pi_{t, s}, Q_{t, s} \rangle \B] \\
    \intertext{because of the sticky setting}
    &=  \sum_s \E_{\{Y_i^s\}} \sum_{i=0}^{N_s} \sum_{t \in (Y^s_i, Y^s_{i+1})} \B[\langle \pi^*_s-\pi_{Y_i^s, s}, Q_{t, s} \rangle \B]
    \intertext{given $|Q_{t+n}(s, a) - Q_t(s, a)| \leq \frac{S+HA}{1-\gamma}n c_T + \frac{2}{\sqrt{T}}$, let $b(n) = \frac{S+HA}{1-\gamma}nc_T$}
    &= \sum_s \E_{\{Y_i^s\}} \sum_{i=0}^{N_s} \sum_{t \in (Y^s_i, Y^s_{i+1})} \B[\langle \pi^*_s-\pi_{Y_i^s, s}, Q_{Y^s_i, s} \rangle + 2b(t-Y^s_i) \B] + \sum_s \sum_{t=1}^T \frac{4}{\sqrt{T}} \\
    \intertext{by the chain rule}
    &= \sum_s \E_{Y^s_{N_s}|Y^s_{N_{s}-1}, \dots, Y^s_0} \dots \E_{Y^s_{1}|Y^s_0} \sum_{i=0}^{N_s} \sum_{t \in (Y^s_i, Y^s_{i+1})} \B[\langle \pi^*_s-\pi_{Y_i^s, s}, Q_{Y^s_i, s} \rangle + 2b(t-Y^s_i) \B] + 4S\sqrt{T}   \\
    %
    \intertext{as the expected length of $[Y_i^s, Y_{i+1}^s)$ is independent of $Y_j^s$ conditioned on $Y_i^s$ for $j > i+1$}
    &= \sum_s \sum_i \E_{Y^s_{i+1}|Y^s_i, \dots, Y^s_0} \sum_{t \in (Y^s_i, Y^s_{i+1})} \B[\langle \pi^*_s-\pi_{Y_i^s, s}, Q_{Y^s_i, s} \rangle + 2b(t-Y^s_i) \B]  + 4S\sqrt{T}
    \intertext{because of the Markov property}
    &= \sum_s \sum_i \E_{Y^s_{i+1}|Y^s_i} \sum_{t \in (Y^s_i, Y^s_{i+1})} \B[\langle \pi^*_s-\pi_{Y_i^s, s}, Q_{Y^s_i, s} \rangle + 2b(t-Y^s_i) \B]  + 4S\sqrt{T}\\
    \intertext{by Lemma~\ref{lemma:expectation}}
    &\leq \sum_s \bigg[ \sum_i \frac{\langle \pi^*_s-\pi_{Y_i^s, s}, Q_{Y^s_i, s} \rangle}{\beta} + \sum_i \E_{Y^s_{i+1}|Y^s_i} \sum_{t \in (Y^s_i, Y^s_{i+1})} 2b(t-Y^s_i) \bigg] + 4S\sqrt{T} \\
    &= \sum_s \bigg[ \sum_i \frac{\langle \pi^*_s-\pi_{Y_i^s, s}, Q_{Y^s_i, s} \rangle}{\beta} + \sum_i \E_{Y^s_{i+1}|Y^s_i} \sum_{t \in (Y^s_i, Y^s_{i+1})} 2 \frac{S+HA}{1-\gamma} (t-Y^s_i) c_T \bigg] + 4S\sqrt{T}\\
    &= \sum_s \bigg[ \sum_i \frac{\langle \pi^*_s-\pi_{Y_i^s, s}, Q_{Y^s_i, s} \rangle}{\beta} + \frac{S+HA}{1-\gamma} \sum_i \E_{Y^s_{i+1}|Y^s_i} (Y^s_{i+1}-Y^s_i)^2 c_T \bigg] + 4S\sqrt{T}
    \intertext{by Lemma~\ref{lemma:variance}}
    &\leq \sum_s \bigg[ \sum_i \frac{\langle \pi^*_s-\pi_{Y_i^s, s}, Q_{Y^s_i, s} \rangle}{\beta} +  \frac{S+HA}{1-\gamma} \sum_i \frac{2}{\beta^3} c_T \bigg]  + 4S\sqrt{T}
    \intertext{$\sum_i \langle \pi^*_s-\pi_{Y_i^s, s}, Q_{Y^s_i, s} \rangle$ is the observed regret at $s$, and summing over $s$ and $i$ results in $T$ steps}
    &= \frac{\tilde{R}^{\text{ob-mdp}}(T)}{\beta} + \frac{2(S+HA)}{\beta^3(1-\gamma)} c_T T + 4S\sqrt{T}
\end{align}
\qed

\section{Proof of Lemma~\ref{lemma:preserve-base}}\label{appendix:slowly-changing}

We first show that the slowly changing property of \local~is preserved by \name.
\begin{lemma}\label{lemma:preserve}
\name~is slowly changing if \local~is slowly changing.
\end{lemma}
{\bf Proof.}
We use $\phi^s$ to denote the \textit{policy} of $\local_s$ (to be distinguished from $\pi$ defined over $\Scal \times \Acal$). The \textit{state space} of $\phi^s$ is the singleton $\{s\}$. We have $||\phi^s_{t+1} - \phi^s_{t}||_{1, \infty} \leq c_T$ because $\local_s$\ is slowly changing.
\begin{align*}
    ||\pi_{t+1} - \pi_{t}||_{1, \infty} &= ||\phi^{s_{t-H}}_{t+1} - \phi^{s_{t-H}}_{t}||_{1, \infty} \leq c_T
\end{align*}
This simply follows the fact that only $\local_{s_{t-H}}$ is updated at time $t$.
\qed

\begin{replemma}{lemma:preserve-base}
While applying Algo.~\ref{alg:bold} as \local, $\tilde{\pi}$ is slowly changing in $\tilde{\mathcal{M}}$, where $\tilde{\pi}_t(a|s \circ h) \coloneqq \pi_t(a|s)$ and $\pi_t$ is produced by \name.
\end{replemma}
\textbf{Proof.} Similar to Lemma~\ref{lemma:preserve}, we use $\phi^s$ to denote the \textit{policy} of $\local_s$. In addition, we use $\phi^{s, h}_t$ to denote the \textit{policy} of $\base^h_s$ of $\local_s$ at time $t$. We have $||\phi^{s, h}_{t+1} - \phi^{s, h}_{t}||_{1, \infty} \leq c_T$, given the slowly changing \base~assumption.
\begin{align}
    ||\tilde{\pi}_{t+H+1} - \tilde{\pi}_{t+H}||_{1, \infty} &= \max_{s\circ h} \sum_a |\tilde{\pi}_{t+H+1}(a|s\circ h) - \tilde{\pi}_{t+H}(a|s\circ h)| \\
    \intertext{by the fact that only $\base^{h_t}_{s_t}$ is updated at $t+H$}
    &= \sum_a |\tilde{\pi}_{t+H+1}(a|s_t \circ h_t) - \tilde{\pi}_{t+H}(a|s_t\circ h_t)| \\
    &= \sum_a |\phi^{s_t, h_t}_{t+H+1} - \phi^{s_t, h_t}_{t+H}| \\
    &= ||\phi^{s_{t}, h_t}_{t+H+1} - \phi^{s_{t}, h_t}_{t+H}||_{1, \infty} \leq c_T
\end{align}
\qed

\section{Proof of Lemma~\ref{lemma:stationary-and-mixing} (Assumptions Hold for $\tilde{\mathcal{M}}$)}\label{appendix:argumented-mdp}

By definition of $\tilde{\mathcal{M}}$, we have

\begin{align}
    \tilde{\pi}(a|s\circ h) &\coloneqq \pi(a|s) 
\end{align}

\begin{equation}
    \tilde{\Pbb}^{\tilde{\pi}} \coloneqq 
    \begin{blockarray}{ccccc}
        & \Scal \circ 1 & \Scal \circ 2 & \cdots & \Scal \circ (H+1) \\
        \begin{block}{c(cccc)}
          \Scal \circ 1 &   & \Pbb^\pi &   &   \\
          \Scal \circ 2 &   &   & \ddots &   \\
          \vdots        &   &   &   & \Pbb^\pi \\
          \Scal \circ (H+1) & \Pbb^\pi &   &   &   \\
        \end{block}
    \end{blockarray}
\end{equation}
where $\Scal \circ h$ stands for concatenating all elements in the set $\Scal$ with $h$.

\bigskip
For brevity, we use $\tilde{\Pbb}^{\pi}$ instead of $\tilde{\Pbb}^{\tilde{\pi}}$ as $\pi$ is sufficient to avoid confusion.

\begin{lemma}\label{lemma:augumented-stationary-distribution}
    (Stationary distribution $\tilde{\mu}^\pi$) For any $\Hcal$-augmented MDP $\tilde{\mathcal{M}}$, if the MDP $\mathcal{M}$ before augmentation satisfies assumption~\ref{asm:fast-mixing}, then there is an unique stationary distribution 
    \begin{equation}
        \tilde{\mu}^\pi = \frac{1}{H+1} \big[ \mu^\pi, \dots, \mu^\pi \big]
    \end{equation}
\end{lemma}
\textbf{Proof.} 

\noindent\textbf{(1) Existence:} 

Let $x_h$ be an row vector (with size $1 \times (H+1)S$ ) that is defined as and ${\bf 0}$ be size $1\times S$
\begin{equation}
    x_h \coloneqq [ \underbrace{{\bf 0}, {\bf 0} \dots}_{\text{$h-1$ zero vectors}}, \mu^\pi \dots, {\bf 0} ]
\end{equation}
Consider a convex combination of $\{ x_h: h=1, \dots, H+1 \}$ multiplied by $\tilde{\Pbb}^\pi$
\begin{align}
    \left( \sum_h \alpha_h x_h \right) \tilde{\Pbb}^\pi &= \sum_h \left( \B[ \underbrace{{\bf 0}, {\bf 0} \dots}_{\text{$h-1$ zero vectors}}, \alpha_h \mu^\pi \dots, {\bf 0} \B] \tilde{\Pbb}^\pi \right)
    \intertext{as density at block $h$ always be pushed to block $h+1$}
    &= \sum_h \left( \B[ \underbrace{{\bf 0}, {\bf 0} \dots}_{\text{$h$ zero vectors}}, \alpha_h \mu^\pi \dots, {\bf 0} \B] \right)
\end{align}

This equality implies that 
\begin{align}
    &\left[ \alpha_1 \mu^\pi, \alpha_2 \mu^\pi, \dots \alpha_{H+1} \mu^\pi \right] = \left[ \alpha_{H+1} \mu^\pi, \alpha_1 \mu^\pi, \dots \alpha_{H} \mu^\pi \right] \\
    \intertext{which implies}
    &\alpha_h = \frac{1}{H+1} \phantom{aaaa} \text{for $h = 1, 2, \dots, H+1$}
\end{align}

Therefore, 
\begin{equation}
    \tilde{\mu}^\pi = \frac{1}{H+1} \big[ \mu^\pi, \dots, \mu^\pi \big]
\end{equation}

\noindent\textbf{(2) Uniqueness: } 

Suppose there exists a row vector $d$ such that $d \tilde{\Pbb}^\pi = d$ and $d \neq \tilde{\mu}^\pi$. Let divide $d$ into $H+1$ blocks as well
\begin{align}
    d = [d_1, d_2, \dots, d_{H+1}]
\end{align}

Multiplying $d$ with the transition kernel $\tilde{\Pbb}^\pi$
\begin{align}
    d \tilde{\Pbb}^\pi &= [d_1, d_2, \dots, d_{H+1}] \tilde{\Pbb}^\pi \\
    &= [d_1, d_2, \dots, d_{H+1}] 
    \begin{pmatrix}
         & \Pbb^\pi & &  \\
         & & \ddots &  \\
         & & & \Pbb^\pi  \\
        \Pbb^\pi & & & 
    \end{pmatrix} \\
    &= [d_{H+1}\Pbb^\pi, d_1\Pbb^\pi, \dots, d_H\Pbb^\pi]
\end{align}
which implies
\begin{align}
    d_h \Pbb^\pi = d_{h+1}
\end{align}
the stationary distribution of $\Pbb^\pi$ is unique, i.e. $\mu^\pi$, and $\|d\|_1 = 1$, which implies
\begin{align}
    d_h = \frac{1}{H+1} \mu^\pi
\end{align}
which contradicts our assumption that $d \neq \tilde{\mu}^\pi$, therefore $\tilde{\mu}^\pi$ is the unique stationary distribution of $\tilde{\Pbb}^\pi$.
\qed\bigskip

\begin{corollary}
    For any $\Hcal$-augmented MDP $\tilde{\mathcal{M}}$, if the MDP $\mathcal{M}$ before augmentation satisfies assumption~\ref{asm:uniform-bound} and assumption~\ref{asm:fast-mixing}, then 
    the stationary distributions $\tilde{\mu}^\pi(s)$ are uniformly bounded away from zero, 
    \begin{equation}
        \inf_{\pi, s}\tilde{\mu}^\pi(s) \geq \frac{\beta}{H+1}\ \ \mathrm{for\ some}\ \beta > 0.
    \end{equation}
\end{corollary}
\textbf{Proof.} Trivially implied by Lemma~\ref{lemma:augumented-stationary-distribution}.
\qed\bigskip

\begin{lemma}
    If assumption~\ref{asm:fast-mixing} holds, then for any two arbitrary distributions $\tilde{d}$ and $\tilde{d}'$ over $\tilde{\Scal}$, we have
    \begin{equation*}
        \sup_\pi\|(\tilde{d}-\tilde{d}')\tilde{\Pbb}^{\pi}\|_1 \leq e^{-1/\tau}\|\tilde{d}-\tilde{d}'\|_1 ,
    \end{equation*}
    where $\tau$ is the same as in assumption~\ref{asm:fast-mixing}.
\end{lemma}
\textbf{Proof.} 

Let $X_{:, j}$ be the $j$-th column of matrix $X$
\begin{align}
    \sup_\pi\|(\tilde{d}-\tilde{d}')\tilde{\Pbb}^{\pi}\|_1 &= \sup_\pi \sum_{s\circ h \in \tilde{\Scal}} \left| (\tilde{d}-\tilde{d}') \tilde{\Pbb}^{\pi}_{:, s\circ h} \right|
    \intertext{As $\tilde{\Pbb}^{\pi}_{:, s\circ h}$ is in the form of $[0, \dots, (\Pbb^\pi_{:, s})^\top, \dots, 0]^\top$, and let $y_h$ be the $h$-th block of row vector $y$, if one divide $y$ into $H+1$ equal blocks (where $H+1$ comes from our construction in Algorithm~\ref{alg:bold})}
    &\leq \sup_\pi \sum_h \sum_s \left| (\tilde{d}-\tilde{d}')_h {\Pbb}^{\pi}_{:, s} \right|
    \intertext{By $\sup_\pi \sum_h \leq \sum_h \sup_\pi$}
    &\leq \sum_h \sup_\pi  \sum_s \left| (\tilde{d}-\tilde{d}')_h {\Pbb}^{\pi}_{:, s} \right|
    \intertext{By assumption~\ref{asm:fast-mixing}}
    &\leq \sum_h e^{-1/\tau} \| (\tilde{d}-\tilde{d}')_h \|_1 \\
    &= e^{-1/\tau} \| \tilde{d}-\tilde{d}' \|_1
\end{align}
\qed 

\section{Proof of Lemma~\ref{lemma:exp3} (\expt\ is Slowly-Changing)}\label{apx:exp3}

We first give Algorithm~\ref{alg:exp3} to provide relevant notations.

\begin{algorithm2e}[H]
\caption{ \expt~}\label{alg:exp3}
\DontPrintSemicolon  

\textbf{Require:}
$\gamma \in [0, 1)$, $\eta_T \in (0, 1], A = |\Acal|$\;

\textbf{Initialize:} $w_1(a) = 1$ for $a \in \Acal$\;

\For {$t = 1, 2, \dots, T$}{
    Set $W_t = \sum_{a=1}^{A}w_t(a)$\; 
    \phantom{Set} $p_t(a) = (1-\eta_T) w_t(a)/W_t + \eta_T/A$\;
    Draw $a_t$ randomly accordingly to $\mathbf{p}_t$\;
    Receive reward $y_t(a_t) \in [0, \tfrac{1}{1-\gamma}]$\;
    For $a=1,2,\dots, A$, set\;
    \vskip -.25in
    \begin{align*}
        \hat{y}_t(a) &= 
        \begin{cases}
            y_t(a)/p_t(a), \text{if~} a = a_t \\
            0, \text{otherwise}
        \end{cases}\\
        w_{t+1}(a) &= w_t(a)\exp((1-\gamma)\eta_T \hat{y}_t(a)/A)
    \end{align*}
    \vspace{-1.5em}
}
\end{algorithm2e}


\begin{replemma}{lemma:exp3}
    \expt~is slowly changing with a rate of $\Ocal(\eta_T/A)$, assuming the feedback $y_t$ is bounded within the range $\left[0, {1}/{(1-\gamma)}\right]$.
\end{replemma}
{\bf Proof.}
We observe it is sufficient to bound $p_{t+1}(a)-p_{t}(a)$ of the action $a$ chosen by the algorithm at time-step $t$. 
We then fix an arbitrary action $a$ to be chosen (and whose weight is updated) and drop it from the notation below w.r.t. $p$, $w$, $\hat{y}$, etc.
\begin{align}
    p_{t+1} - p_{t}
    = & \  (1-\eta_T) \left( \frac{w_{t+1}}{W_{t+1}} - \frac{w_t}{W_t} \right)  \\
    =& \ (1-\eta_T) \left( \frac{ w_te^{(1-\gamma)\eta_T \hat{y}_t / A}}{W_t +
    w_t (e^{\eta_T \hat{y}_t / A} - 1) } -
    \frac{w_t}{W_t} \right)  \\
    \le & (1-\eta_T) \left( \  \frac{w_t e^{(1-\gamma)\eta_T \hat{y}_t / A} }{W_t} -
    \frac{w_t}{W_t} \right)  \\
   =& \ (1-\eta_T) \left( \frac{w_t(e^{(1-\gamma)\eta_T \hat{y}_t / A}-1) }{W_t} \right)  \\
   \le & \ (1-\eta_T) \left(2(1-\gamma) \left(\frac{ \eta_T \hat{y}_t}{A}\right) \left(\frac{w_t}{W_t}\right) \right) \label{eq:exminonbbd} \\
   \le & (1-\eta_T) \left( 2\left(\frac{ \eta_T }{A p_t}\right) \left(\frac{w_t}{W_t}\right) \right)   \\
   \le &  2{\eta_T}/{A} \label{eq:wtsub}. 
\end{align}
\eqref{eq:exminonbbd} follows from that
$e^x - 1 < 2x$ for $0 \le x \le 1$ and \eqref{eq:wtsub} follows from
$p_t \ge (1-\eta_T)\left( w_t/ W_t\right)$.

As mentioned in Section~\ref{sec:exp3}, to achieve $\tilde{\Ocal}(\sqrt{AT})$ regret, \expt~is run with a learning rate of {$\eta_T = 
\tilde{\Ocal}{(\sqrt{A/T})}$}, which means it is slowly changing with a rate of $c_T = \tilde{\Ocal}(\sqrt{1/AT})$.

\end{document}